%% file: main.tex
\documentclass[lettersize,journal]{IEEEtran}
\usepackage{amsmath,amsfonts}
\usepackage{algorithmic}
\usepackage{algorithm}
\usepackage{array}
\usepackage[caption=false,font=footnotesize,labelfont=sf,textfont=sf]{subfig}
\usepackage{textcomp}
\usepackage{stfloats}
\usepackage{url}
\usepackage{verbatim}
\usepackage{graphicx}
\usepackage{cite}
\usepackage{makecell}
\usepackage{multirow}
\usepackage{enumitem}
\begin{document}

\title{Perceptual Video Coding for Machines via Satisfied Machine Ratio Modeling}

\author{
  Qi~Zhang,~\IEEEmembership{Student Member,~IEEE,}
  Shanshe~Wang,~\IEEEmembership{Member,~IEEE,}
  Xinfeng~Zhang,~\IEEEmembership{Senior~Member,~IEEE,}
  Chuanmin~Jia,~\IEEEmembership{Member,~IEEE,}
  Zhao~Wang,~\IEEEmembership{Member,~IEEE,}
  Siwei~Ma,~\IEEEmembership{Fellow,~IEEE,}
  and~Wen~Gao,~\IEEEmembership{Fellow,~IEEE}
  \thanks{
    Corresponding author: Siwei Ma (Email: swma@pku.edu.cn) and Xinfeng Zhang (Email: xfzhang@ucas.ac.cn).
  }
  \thanks{
    Qi Zhang, Shanshe Wang, Zhao Wang, Siwei Ma, and Wen Gao are with National Engineering Research Center of Visual Technology, School of Computer Science, Peking University, Beijing, China. (Emails: qizhang@stu.pku.edu.cn, \{sswang, zhaowang, swma, wgao\}@pku.edu.cn)
  }
  \thanks{
    Xinfeng Zhang is with University of Chinese Academy of Sciences, Beijing, China. Chuanmin Jia is with Wangxuan Institute of Computer Technology, Peking University, Beijing, China. (Email: cmjia@pku.edu.cn). Siwei Ma and Wen Gao are also with Peng Cheng Laboratory, Shenzhen, China.
  }
}

\markboth{Journal of \LaTeX\ Class Files,~Vol.~1, No.~1, January~2024}%
{Shell \MakeLowercase{\textit{et al.}}: A Sample Article Using IEEEtran.cls for IEEE Journals}


\maketitle

\begin{abstract}
  Video Coding for Machines (VCM) aims to compress visual signals for machine analysis. However, existing methods only consider a few machines, neglecting the majority. Moreover, the machine's perceptual characteristics are not leveraged effectively, resulting in suboptimal compression efficiency. To overcome these limitations, this paper introduces Satisfied Machine Ratio (SMR), a metric that statistically evaluate the perceptual quality of compressed images and videos for machines by aggregating satisfaction scores from them. Each score is derived from machine perceptual differences between original and compressed images. Targeting image classification and object detection tasks, we build two representative machine libraries for SMR annotation and create a large-scale SMR dataset to facilitate SMR studies. We then propose an SMR prediction model based on the correlation between deep feature differences and SMR. Furthermore, we introduce an auxiliary task to increase the prediction accuracy by predicting the SMR difference between two images in different quality. Extensive experiments demonstrate that SMR models significantly improve compression performance for machines and exhibit robust generalizability on unseen machines, codecs, datasets, and frame types. SMR enables perceptual coding for machines and propels VCM from specificity to generality. Code is available at \url{https://github.com/ywwynm/SMR}.
\end{abstract}

\begin{IEEEkeywords}
  Video Coding for Machines, Perceptual Coding, Just Noticeable Difference, Satisfied User Ratio
\end{IEEEkeywords}

\section{Introduction} \label{sec:introduction}

\IEEEPARstart{T}{he} use of intelligent machines has surged in recent years, with many designed for analyzing and understanding visual data like images and videos. Over the past decade, advancements in AI techniques, including convolutional neural network \cite{krizhevsky2012imagenet}, deep learning \cite{lecun2015deep}, residual \cite{he2016deep} block, knowledge distillation \cite{hinton2015distilling}, network architecture searching \cite{tan2019mnasnet}, attention mechanism \cite{dosovitskiy2020image}, self-supervised learning \cite{he2022masked}, \textit{etc.}, have significantly enhanced the capability and efficiency of machine vision systems (MVS). Presently, machines exceed human performance in various visual analysis tasks \cite{he2015delving,silver2016mastering} and are increasingly adopted in diverse applications.

Like humans seeking high-quality images for visual satisfaction, machines pursue high-quality visual data for accurate analysis. This need is typically met during training phases, where machines learn from high-quality image datasets like ImageNet \cite{deng2009imagenet} and Microsoft COCO \cite{lin2014microsoft}. However, in real-world scenarios, image quality can be compromised by blurriness, noise, poor luminance, \textit{etc.}, challenging machines to extract accurate features for visual analysis \cite{dodge2016understanding}. The most common distortion is compression artifacts, as images and videos need to be compressed to reduce transmission and storage costs. Several studies \cite{ma2018joint,zhang2021just} have revealed that compression can significantly decrease the accuracy of different machines and tasks, particularly at low bit-rates. Therefore, a great challenge is posed for image and video coding to ensure the effectiveness and reliability of machines in different applications, where compression quality varies to meet real-world demands.

In the past decades, image and video coding standards such as JPEG \cite{wallace1991jpeg}, AVC \cite{wiegand2003overview}, HEVC \cite{sullivan2012overview}, VVC \cite{bross2021overview}, and AVS3 \cite{zhang2019recent} have greatly improved compression efficiency. The emerging neural codecs \cite{ma2019image} also offer novel and encouraging solutions. However, they primarily optimize for human vision systems (HVS) to enhance visual fidelity and quality at comparable bit-rates. Recently, several MVS-oriented compression methods have been explored to fill this gap. Some works propose to modify current codecs, such as refining parameter selection \cite{liu2017recognizable,li2021task}, rate-distortion optimization \cite{zhang2016joint,zhang2020novel}, and bit allocation \cite{choi2018high,huang2021visual} modules. Another direction is to learn end-to-end compression networks with machine vision task-specific constraints \cite{luo2018deepsic,torfason2018towards,bai2022towards}. In line with these explorations, the standardization progress has also been started by MPEG, namely Video Coding for Machines (VCM) \cite{gao2021recent}, aiming to establish a promising platform for advancing MVS-oriented coding techniques and promoting the deployment of machine vision applications in various domains.

However, existing MVS-oriented compression methods suffer from two notable drawbacks. First, they are designed and evaluated with limited machines considered, thus lacking the generalizability. For instance, the Common Test Conditions (CTC) during MPEG VCM standardization \cite{vcm-ctc} assess compression performance using only a single machine for each task, introducing inherent bias in the evaluation and making the results more subjective (accurate only for that specific machine) than objective (generalized enough for most machines). Such bias becomes more problematic in the real world where machine information is unavailable during compression, or when machines are updated or replaced due to the diverse application requirements. Second, the MVS characteristics are underutilized, particularly the machine perceptual characteristics on distinguishing between good and bad compression quality, limiting compression performance. This compression inefficiency leads to additional costs in transferring, storing, and processing the input visual data for machines, which becomes increasingly problematic in the era of AI due to the gigantic data volume and ubiquitous applications.

Revisiting HVS-oriented compression methods, perceptual coding is a valuable reference addressing these drawbacks. Specifically, HVS has inherent limitations in perceiving image quality differences only beyond a certain threshold \cite{zhang2023survey}. Such characteristics bring in perceptual redundancy that can be removed during compression. To capitalize on this, the Just Noticeable Difference or Just Noticeable Distortion (JND) concept is introduced \cite{jayant1993signal} and modeled \cite{chou1995perceptually,yang2005just,liu2019deep,shen2020just,jiang2022toward}. JND identifies the minimum compression quality degradation that HVS can perceive, establishing an optimal compression operation point. As visual acuity varies among human individuals, JND is further extended to Satisfied User Ratio (SUR) \cite{wang2017videoset,fan2019net,zhang2020satisfied,zhang2021deep}. SUR is derived from the cumulative distribution of JND locations obtained from a large population of human subjects, preventing inaccurate or insufficient perceptual characteristics capturing from subject bias. Therefore, SUR measures image quality in a more generalizable manner. Unlike a singlular JND point, SUR forms a continuous and deterministic model that provides multiple operation points for compression, improving its versatility and applicability.

JND and SUR successfully model the HVS behaviors, enabling the HVS-oriented perceptual coding. More recently, a few studies demonstrate the existence of JND for machines \cite{zhang2021just,jin2021just,zhang2023learning} and highlight its potential in MVS-oriented perceptual coding. However, unifying the compression performance and generalizability simultaneously for MVS in a practical manner remains an issue of paramount significance. In this work, we make the initial attempt to study SUR for machines to tackle this unprecedented challenge. Our contributions are presented as follows.

\begin{itemize}[leftmargin=10pt]
  \item We demonstrate the necessity of involving diverse machines in VCM through two pilot studies. The first one reveals that different machines have distinct perceptions on images in the same compression quality. The second one indicates that optimizing the encoder for one machine can decrease the performance of another. To our knowledge, we are the first to comprehensively address machine diversity in VCM.
  \item We propose a novel concept, Satisfied Machine Ratio (SMR), to model the general MVS characteristics for VCM. SMR is defined as the proportion of machines with higher satisfaction scores on a compressed image or video frame above a reasonable threshold. Each satisfaction score is calculated based on the machine perceptions' differences on original and compressed images. To our knowledge, we are the first to explore SUR for machines.
  \item We study SMR on two fundamental machine vision tasks: image classification and object detection. We build two representative machine libraries to analyze MVS behaviors, comprising up to 72 and 98 diverse machines for each task, respectively. Utilizing these machines for SMR annotation, we create a large-scale SMR dataset containing over 27 million images in 37 compression quality levels and more than 593 million ground truth labels. This dataset facilitates further SMR studies.
  \item We analyze the SMR dataset to uncover MVS characteristics individually at the image level and aggregately at the dataset level. We find that MVS exhibits unique JND characteristics on many images when facing compression quality degradations, diverging from the patterns seen in HVS. We also find that SMR and SUR have substantial differences, highlighting the distinct perceptual characteristics of machines compared to humans.
  \item We propose the SMR modeling task. We discover a nonlinear negative correlation between deep feature differences and SMR. Leveraging this correlation, we design a full-reference SMR prediction model to predict the SMR of any image or video frame. This model serves as a solid baseline for the SMR prediction and SMR-guided coding optimization task. Furthermore, we introduce an auxiliary task of predicting the SMR difference between a pair of images in different compression quality, which fully utilizes all labeled data and increases the SMR prediction accuracy.
  \item We conduct extensive experiments to validate the effectiveness of our proposed SMR models in predicting SMR and improving the compression performance for machines. We achieve significant coding gains by using the predicted SMR as the optimization target of codecs. Crucially, our SMR models have strong generalizability on unseen machines, codecs, datasets, and frame types. These evaluations establish a reliable benchmark for future works in this area.
\end{itemize}

In conclusion, SMR enables perceptual coding for machines and propels VCM from specificity to generality. Furthermore, considering many diverse machines instead of one or a few specific machines can benefit other machine vision and image processing areas, including image and video quality assessment and enhancement, adversarial attack, privacy-protected visual analysis, and beyond. 

The rest of our paper is organized as follows. Section \ref{sec:related works} presents several related works. Section \ref{sec:machine diversity} reveals machine diversity and its potential negative impact on existing codecs. In Section \ref{sec:smr}, the concept of SMR is proposed and defined in detail. Subsequently, focusing on two fundamental vision tasks, we construct two machine libraries for SMR annotation and build a large-scale SMR dataset in Section \ref{sec:smr dataset}. In Section \ref{sec:smr modeling}, two deep learning-based SMR models are proposed for SMR prediction. Section \ref{sec:experiments} shows the experimental results of evaluating the SMR model's performance and generalizability. Finally, Section \ref{sec:conclusion} concludes the paper.


\section{Related Works} \label{sec:related works}

\subsection{Video Coding for Machines} \label{subsec:vcm}

VCM aims at compressing visual signals efficiently while keeping their utility for machine analysis. Several works are proposed to improve the coding performance for MVS in multiple ways \cite{yang2021video}. The first it to optimize conventional image and video codecs for MVS. Liu \textit{et al.} \cite{liu2017recognizable} trained a support vector machine (SVM) to predict whether the image quality is good enough for correct machine analysis by the differences of SIFT features before and after compression. The prediction is then used for choosing adequate coding configurations for MVS. Li \textit{et al.} \cite{li2021task} adopted deep reinforcement learning technique to select quantization parameters for coding tree units (CTU) in HEVC, where the reward function is related to both bit-rate and machine analysis accuracy. Zhang \textit{et al.} \cite{zhang2020novel} introduced task-specific quality metric in the rate-distortion optimization (RDO) procedure of HEVC to obtain better analysis results under the same compression rate. In \cite{choi2018high}, a bit allocation strategy is designed to assign more bits to regions with higher activations from shallow task network layers. Similarly, Huang \textit{et al.} \cite{huang2021visual} proposed a multi-scare feature distortion metric for RDO with salient areas detected by a pre-trained region proposal network.

The second way is to train an end-to-end neural codec for MVS. In \cite{luo2018deepsic}, two types of deep semantic image coding frameworks were proposed, distinguished by analysis being performed on encoder or decoder side. The codec combined bit-rate, signal-level, and task-level errors to form the training loss function. Torfason \textit{et al.} \cite{torfason2018towards} trained a network for both compression and analysis, where the compact representation generated by the codec is directly used for analysis without reconstructing image textures. Codevilla \textit{et al.} \cite{codevilla2021learned} and Chamain \textit{et al.} \cite{chamain2021end} attached various task networks to produce different losses for optimizing the codec and improving its generalizability. To further increase the coding efficiency, state-of-the-art AI techniques like vision transformers \cite{bai2022towards} and contrastive learning \cite{dubois2021lossy} are introduced. More recently, scalable layered coding becomes a trending compression framework for both humans and machines. Akbari \textit{et al.} \cite{akbari2019dsslic} proposed to include a semantic segmentation map as the base layer, which is compressed losslessly for analysis tasks. Their framework also contains a primary enhancement layer where textures are generated by a decoder network, and the residual between the original image and generated one is also compressed as the secondary enhancement layer. The segmentation map was replaced by an edge map in \cite{hu2020towards,chang2023semantic} or deep features in \cite{wang2019scalable,wang2021towards}. Yan \textit{et al.} \cite{yan2020semantically} performed layered coding to features, where features from different layers of the network have distinct semantic granularities for multiple analysis tasks and texture reconstruction as well \cite{yan2021sssic}. Similarly, Liu \textit{et al.} \cite{liu2021semantics} designed a transformation network based on a lifting scheme to convert images into multi-level scalable representations for end-to-end compression. Tu \textit{et al.} \cite{tu2021semantic} utilized the correlation between neighboring network layers and the different importance between object and background areas to improve compression performance. Choi \textit{et al.} \cite{choi2021latent} purposefully split the latent representations into two parts during training, where the first part stores semantics for analysis as the base layer and the second stores textures information for reconstruction as the enhancement layer. The generated scalable bit-stream can be arranged structurally, where different parts of the bit-stream stores information from different layers \cite{sun2020semantic,choi2022scalable} to support on-demand transmission and storage.

Many efforts are also made from standardization perspective. As early as in the late 1990s, MPEG has explored the content description of images and videos for intelligent identification, categorization, and user-browsing, from which the international MPEG-7 visual standard was born \cite{sikora2001mpeg}. In 2015, the Compact Descriptor for Visual Search (CDVS) standard \cite{duan2015overview} was declared, which defines the bit-stream of descriptors (\textit{i.e.} hand-crafted visual features) and the descriptor extraction process for image matching and retrieval task. Its successor, the Compact Descriptor for Video Analysis (CDVA) \cite{duan2018compact}, was finalized in 2019, which introduces deep features and enables machine analysis on videos. These former standards promote the standardization of Digital Retina \cite{gao2021digital} in China, which consists of three bit-streams (videos, features, and models) to support applications at scale. Meanwhile, internationally, MPEG began to develop the Video Coding for Machines standard \cite{gao2021recent}, pursuing high-efficiency visual signal compression, better machine analysis performance, computational offloading, and privacy preserving at the same time.

Although there is significant progress for VCM, existing methods have the drawbacks mentioned in Section \ref{sec:introduction}. In this work, we will address these issues to improve VCM for satisfying the compression needs of diverse machines efficiently.

\subsection{Perceptual Coding for Humans}

HVS cannot perceive tiny distortions caused by compression. JND models take advantage of this behavior to determine an optimal operation point for compression that balances bit-rate and perceptual quality. To study JND, several image and video databases are created such as MCL-JCI \cite{jin2016statistical} and VideoSet \cite{wang2017videoset}. JND can be modeled in different granularities like pixel, sub-band, and picture or video wise \cite{zhang2023survey}. Picture or video wise JND is more reasonable as HVS perceives image/video entirely instead of pixel or sub-band individually, and is also more applicable because it can be used more easily by adjusting coding parameters. Huang \textit{et al.} \cite{huang2017measure} predicted JND with a support vector regressor (SVR) using hand-crafted spatio-temporal features. Liu \textit{et al.} \cite{liu2019deep} developed a deep learning-based JND estimation framework, converting the task into a binary classification problem to predict whether two images are perceptually identical. Tian \textit{et al.} \cite{tian2020just} trained a CNN to predict the JND of a distorted image without the original image as reference, and also used an SVR to predict the number of JND levels. In \cite{shen2020just}, the structural visibility was firstly predicted at the image-patch level, and then the image-level JND can be estimated by collecting results from all patches. Since humans are different, JND is extended to SUR to represent the HVS characteristics of many human subjects. Wang \textit{et al.} \cite{wang2018prediction} predicted SUR with quality degradation and temporal masking features. Fan \textit{et al.} \cite{fan2019net} designed a Siamese CNN with shared weights to predict SUR based on transfer learning. The network architecture is optimized in \cite{lin2020featnet} by fusing features from different layers to decrease the SUR estimation error. Zhang \textit{et al.} \cite{zhang2020satisfied} combined saliency distribution, masking effect, quality degradation, and bit-rate change features to train an SVR for predicting SUR. In \cite{zhang2021deep}, a two-stream CNN was developed to predict video-wise SUR using spatio-temporal features, which are fused at the quality score and feature levels. Due to the differences in principle and target between HVS and MVS, existing perceptual coding methods for humans cannot be directly applied for machines, which will be studied extensively in this work.

\subsection{Perceptual Coding for Machines}

It is still at an early age for perceptual coding for machines. There are a few works exploring JND for MVS. Zhang \textit{et al.} \cite{zhang2021just} is the first to propose Just Recognizable Distortion (JRD), which presents the maximum compression distortion that will not cause machine recognition failure. They built a large JRD dataset where JRD is measured by the quantization parameter used for coding, and developed an ensemble learning-based JRD prediction framework that can improve the task performance under the same bit-rate. Zhang \textit{et al.} \cite{zhang2023learning} further improved the JRD prediction performance by introducing an error-tolerance strategy. Jin \textit{et al.} \cite{jin2021just} demonstrated the existence of JND for MVS and proposed a generative model to predict pixel-level JND map. According to their experiments, machines can tolerate the JND image with an average peak signal-to-noise ratio (PSNR) of only $9.56$dB. However, these works did not consider machine diversity thoroughly and had some other problems. In \cite{zhang2021just}, the JRD of four machines for two analysis tasks are studied individually. In \cite{zhang2023learning}, just one machine is used. In \cite{jin2021just}, only four old-fashioned machines for the single image classification task are considered together, and the JND model cannot be directly used in VCM codecs. In this work, we will further improve the generalizability for perceptual coding for machines practically.

\section{Machine Diversity} \label{sec:machine diversity}

We first study machine diversity through two experiments. We carefully select 12 different machines for these studies, which are VGG-19 \cite{simonyan2014very}, ResNet-50, ResNet-101 \cite{he2016deep}, ResNeXt-101 \cite{xie2017aggregated}, DenseNet-161 \cite{huang2017densely}, MobileNet-v3-Large \cite{howard2019searching}, EfficientNet-B0, EfficientNet-B4 \cite{tan2019efficientnet}, Vision Transformer-B/16 \cite{dosovitskiy2020image}, ConvNeXt-Base \cite{liu2022convnet}, Swin-T, and Swin-B \cite{liu2021swin}. These machines not only vary in their macro architectures, such as CNNs \textit{vs.} transformers, but also in their internal modules (VGG \textit{vs.} DenseNet), size or complexity (MobileNet \textit{vs.} the others), depth (ResNet-50 \textit{vs.} ResNet-101), overall scale (EfficientNet-B0 \textit{vs.} EfficientNet-B4), and degree of ``modernization'' \cite{liu2022convnet} (ResNet-50 \textit{vs.} ConvNeXt-Base). Given their widespread use in various vision tasks and real-world applications, these machines provide a robust basis for a comprehensive examination of diversity.

\textbf{In the first experiment, we demonstrate that different machines have distinct MVS characteristics by comparing their perceptions under the same compression quality.} We compress images from the val2017 dataset of MS COCO \cite{lin2014microsoft} with HM-16.24 \cite{hm}, the reference software of HEVC. Each image is compressed as an intra video frame. Various quality levels are produced by adjusting the quantization parameter (QP) for compression. Specifically, 20 QPs are selected, which are $32, 33, \dots, \text{and } 51$, where a higher QP indicates worse quality. We employ relatively high QPs in this experiment as lower QPs typically result in tiny distortions that are often negligible for MVS \cite{ma2018joint,zhang2021just}.

We randomly crop $10,000$ objects from COCO images with ground truth bounding boxes and obtain the perceptions of the selected machines on them. Initially, all machines have the weights from PyTorch \cite{paszke2019pytorch}, trained on ImageNet, and subsequently fine-tuned on the original COCO train2017 dataset using object images to aligh with COCO semantics. Since machines are created for image content analysis, their perceptions are equivalent to analysis results. Therefore, for each machine $M_m$ and each object image $I_i$, we record the top-1 predictions (\textit{i.e.} the category of the highest probability) across 21 quality levels, \textit{i.e.} compressed by 20 QPs plus the uncompressed one. The perceptions are denoted as $M_m(I_i^{32}), M_m(I_i^{33}), \dots, M_m(I_i^{51}), \text{and } M_m(I_i^{0})$. We then calculate a \textit{\textbf{machine perception consistency label}} for the perception of $M_m$ to $I_i$ in compression quality $q_k$ using the following formulation:

\begin{equation} \label{eq:machine perception consistency label}
  L(M_m; I_i^{q_k}) = 
  \begin{cases}
    1, & \text{if}\ M_m(I_i^{q_k}) = M_m(I_i^0) \\
    0, & \text{otherwise}
  \end{cases}.
\end{equation}

\noindent When the label is 1, the machine's perception on the compressed image remain unchanged from the uncompressed version, suggesting no impact from compression. Otherwise, the compression has a substantial impact on the machine. For machine $M_m$, we generate a \textit{\textbf{machine perception consistency sequence}} of such labels on object image $I_i$ across all compression quality levels, which is denoted as 

\begin{equation}
  \text{Seq}(M_m; I_i) = [L(M_m; I_i^{32}), L(M_m; I_i^{33}), \dots, L(M_m; I_i^{51})].
\end{equation}

\noindent Finally, we calculate the \textit{\textbf{diversity score}} for a pair of machines $M_m, M_n$ on $I_i$ by 

\begin{equation}
  S_{\text{div}} = \text{HD}(\text{Seq}(M_m; I_i), \text{Seq}(M_n; I_i)),
\end{equation}

\noindent where $\text{HD}(\cdot)$ returns the Hamming distance of two sequences. Therefore, a high diversity score between two machines means that their perceptions are significantly different on the same compression quality of an image.

\begin{figure}[htbp]
  \centering
  \includegraphics[width=0.42\textwidth]{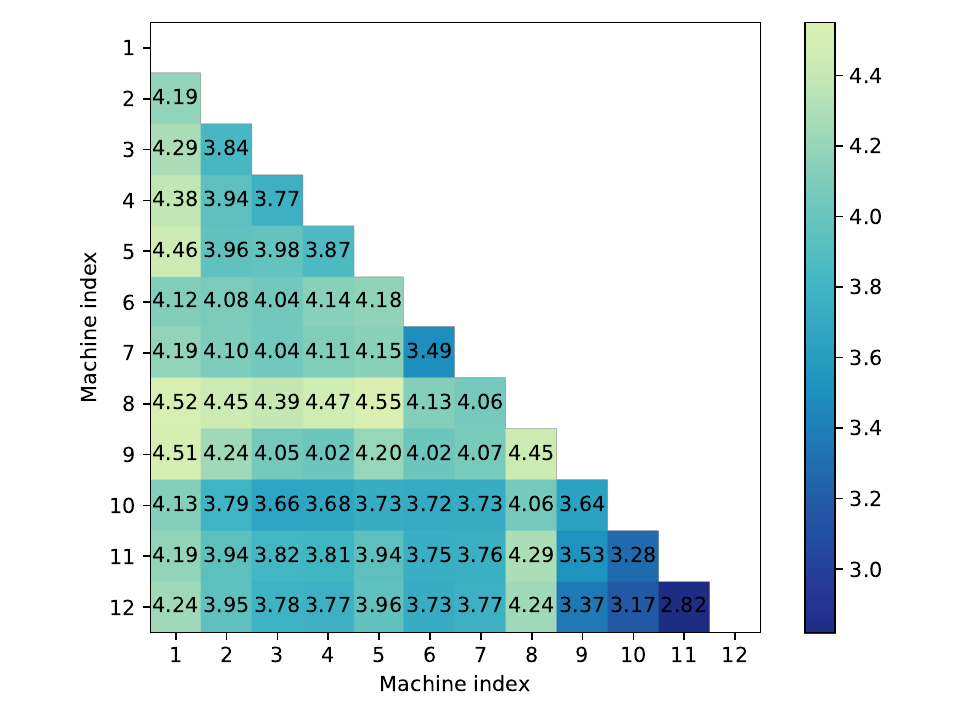}
  
  \caption{The diversity scores of pairs of machines measured by the average Hamming distance of machine perception consistency sequences.}
  \label{fig:machine pair diversity score}
\end{figure}

The diversity scores between pairs of the selected machines are depicted in Fig.~\ref{fig:machine pair diversity score}, where the results are averaged from $10,000$ object images three times. For each time, the objects are re-selected. The axis numbers correspond to the indices of different machines, which are maintained in the same order as their appearances in the first paragraph of this section. The average overall diversity score across all machine pairs is $3.98$, with a standard deviation of $0.33$, indicating that machines' perceptions differ in $19.9\%$ of all 20 compression quality levels. Among specific machine pairs, the three highest diversity scores are observed between DenseNet-161 and EfficientNet-B4 ($4.55$), VGG-19 and EfficientNet-B4 ($4.52$), and VGG-19 and Vision Transformer-B/16 ($4.51$). Conversely, the three lowest scores are found between Swin-T and Swin-B ($2.82$), ConvNeXt-Base and Swin-B ($3.17$), and ConvNeXt-Base and Swin-T ($3.28$). Notably, machine diversity is consistently non-negligible, even among machines with the same macro architecture but only differing in depth or overall scale, such as ResNet-50 and ResNet-101 ($3.84$) or EfficientNet-B0 and EfficientNet-B4 ($4.06$). Moreover, machine diversity exists commonly on most images. Specifically, when considering all $\tbinom{12}{2}=66$ machine pairs, let $P_I$ represent the proportion of images where $S_{\text{div}}>0$ for at least $P_M$ machine pairs, then $P_I=[91.65\%, 82.90\%, 76.98\%, 65.54\%]$ when $P_M=[1, 33(50\%), 50(75\%), 59(90\%)]$, respectively.

How does this diversity impact VCM? \textbf{We demonstrate that a VCM codec optimized for one machine may be ineffective or even detrimental for another.} In this study, we treat the codec as a black box, randomly modify the coding configuration, and check whether these modifications bring consistent changes to different machines' perceptions. Specifically, for object image $I_i$, we first select a random QP in $[32, 51]$, namely $q_\text{base}$, and obtain the machine perception consistency label $L_m^\text{base}=L(M_m; I_i^{q_\text{base}})$ for machine $M_m$. Next, we randomly modify $q_\text{base}$ by adding or subtracting a value in $[1, 5]$, resulting in a modified QP as $q_\text{mod}$ and record the new label $L_m^\text{mod}$. The modified QP will be adjusted to $[32, 51]$ if it exceeds the QP boundary. It is also ensured that $q_\text{mod} \neq q_\text{base}$. Finally, we compare the changes of labels of a pair of machines $(M_m,M_n)$, which are ${\Delta}L_m=L_m^\text{mod}-L_m^\text{base}$ and ${\Delta}L_n=L_n^\text{mod}-L_n^\text{base}$. If ${\Delta}L_m={\Delta}L_n$, the modification is considered as \textit{\textbf{satisfying}} as machines' perceptions are affected in the same direction. Otherwise, we consider the following cases (note that $m/n$ in ${\Delta}L_{m/n}$ means $m$ or $n$, and $n/m$ in $L_{n/m}^\text{base}$ and $L_{n/m}^\text{mod}$ means $n$ or $m$, their orders are corresponding and intentional):

\begin{enumerate}[leftmargin=14pt, label={\arabic*.}]
  \item ${\Delta}L_m\times{\Delta}L_n=-1$: \textit{\textbf{unsatisfying}} as one machine's perception improves while another deteriorates, \textit{i.e.}, the codec optimization for one machine leads to worse performance for another.
  \item ${\Delta}L_{m/n}=-1 \land L_{n/m}^\text{base}=L_{n/m}^\text{mod}=0$: \textit{\textbf{irrelevant}} as the modification is not optimization that should be withdrawn.
  \item ${\Delta}L_{m/n}=-1 \land L_{n/m}^\text{base}=L_{n/m}^\text{mod}=1$: \textit{\textbf{unsatisfying}} when $q_\text{mod}>q_\text{base}$ as one machine's perception remains consistent after bit-rate decrease while another deteriorates, \textit{i.e.}, the codec optimization for one machine leads to worse performance for another. But it is \textit{\textbf{irrelevant}} like case 2 when $q_\text{mod}<q_\text{base}$.
  \item ${\Delta}L_{m/n}=1 \land L_{n/m}^\text{base}=L_{n/m}^\text{mod}=0$: \textit{\textbf{unsatisfying}} as one machine's perception improves while another remains inconsistent, \textit{i.e.}, the codec optimization for one machine is ineffective for another.
  \item ${\Delta}L_{m/n}=1 \land L_{n/m}^\text{base}=L_{n/m}^\text{mod}=1$: \textit{\textbf{satisfying}} as one machine's perception improves and another remains consistent.
\end{enumerate}

\begin{figure}[htbp]
  \centering
  \includegraphics[width=0.48\textwidth]{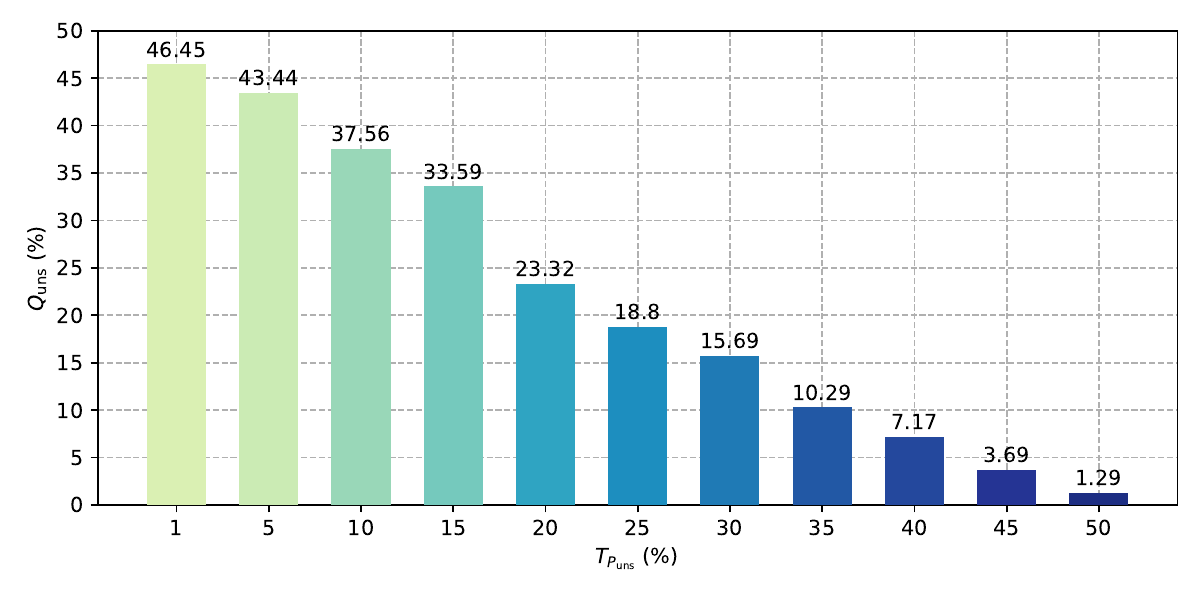}
  \caption{Numbers of unsatisfying codec modifications in percentage under different $T_{P_\text{uns}}$ values.}
  \label{fig:unsatisfying codec modifications}
\end{figure}

To make the investigation more reliable, we also randomly select $10,000$ object images and repeat the experiment three times. For each modification, we record the proportion of unsatisfying cases as $P_\text{uns}$ across all 66 machine pairs. Then, the modification itself is deemed as unsatisfying when $P_\text{uns} \geq T_{P_\text{uns}}$, where $T_{P_\text{uns}}$ is a threshold expressed in percentage. The proportions of unsatisfying modifications (noted as $Q_\text{uns}$) under different $T_{P_\text{uns}}$ values are illustrated in Fig.~\ref{fig:unsatisfying codec modifications}. Due to machine diversity, a codec modification aimed at optimizing for one machine can easily bring unsatisfying results for another. Specifically, up to $46.45\%$ of modifications are unsatisfying when $T_{P_\text{uns}}=1\%$. In a more practical scenario where $T_{P_\text{uns}}=20\%$, $23.32\%$ of modifications are unsatisfying. As $T_{P_\text{uns}}$ increases, $Q_\text{uns}$ decreases as expected because there are also many satisfying modifications. Among all modifications, unsatisfying cases 1, 3, and 4 occur with frequencies of $22.19\%$, $21.51\%$, and $32.01\%$, respectively, indicating no significant disparity in their occurrences. In conclusion, we should carefully consider machine diversity in VCM.

\section{Satisfied Machine Ratio} \label{sec:smr}

To properly address machine diversity in VCM and facilitate MVS-oriented perceptual coding, we propose the concept of \textit{\textbf{SMR}}. The definition of SMR is given as follows. Let $I_0$ be an original image or video frame compressed into several distorted variants $I_{q_1}, I_{q_2}, \ldots, I_{q_n}$, where $q_i$ denotes a specific quality that deteriorates as $i$ increases. Each machine $M_j$ in the machine library $\mathbb{M}$ assigns a \textit{\textbf{satisfaction score}} to $I_{q_i}$, represented by $\text{S}(M_j; I_{q_i})$ and will be described in detail soon. Then, the SMR of $I_{q_i}$ is calculated as

\begin{equation} \label{eq:smr}
  \text{SMR}(I_{q_i}) = \frac{| \{M_j \mid \text{S}(M_j; I_{q_i}) \geq T_S\} |}{|\mathbb{M}|},
\end{equation}

\noindent where $T_S$ is a threshold for satisfaction score, and $|\cdot|$ counts the number of elements in the set. As such, SMR statistically quantifies the proportion of machines satisfied with the image compression quality. During the SMR calculation, $T_S$ controls the strictness of how machine perception consistency is judged. A high $T_S$ indicates that the machine is satisfied with the compression quality only when its perception is very similar between the original and compressed image, typically indicating high analysis accuracy. Conversely, a low $T_S$ allows for a greater machine perception variation.

The calculation of satisfaction scores differs across different machine vision tasks due to their unique objectives and evaluation metrics. In this study, we focus on SMR for two fundamental tasks: image classification and object detection. For image classification, the satisfaction score function is defined in a similar form as Eq. \eqref{eq:machine perception consistency label} in Section \ref{sec:machine diversity}, but is extended for more flexibility:

\begin{equation} \label{eq:satisfaction score function for cls}
  \text{S}_K(M_j; I_{q_i}) = 
  \begin{cases}
    1, & \text{if}\ \text{Top-1}(M_j(I_{q_i})) \ \subseteq \ \text{Top-}K(M_j(I_0)) \\
    0, & \text{otherwise}
  \end{cases},
\end{equation}

\noindent where $M_j(I_{q_i})$ is the output probabilities for each class from $M_j$ given $I_{q_i}$ as input, and $\text{Top-1}(\cdot)$ and $\text{Top-}K(\cdot)$ return catgory indices with top-1 and top-K probability value(s), respectively. When calculating SMR for image classification using Eq. \eqref{eq:smr}, $T_S=1$. Therefore, SMR for image classification represents the proportion of machines with consistent class predictions on original and compressed images.

The satisfaction score function for object detection is more complicated, considering various performance evaluation factors like the intersection over union (IOU) between the detected bounding boxes and ground truth, class confidence of a detected object, and the number of detections allowed for the evaluation. The mean average precision (mAP) is the \textit{de facto} metric for this task that incorporates these factors. We thus define the satisfaction score function for object detection using the mAP metric as follows

\begin{equation} \label{eq:satisfaction score function for det}
  \text{S}_\mathbb{T}(M_j; I_{q_i}) = \text{mAP}_{T_\text{IOU}}(M_j(I_{q_i}), \mathcal{F}_{T_\text{conf}}(M_j(I_0))),
\end{equation}

\noindent where $T_\text{IOU}$ is the threshold value of IOU when calculating mAP, and $\mathcal{F}_{T_\text{conf}}(\cdot)$ is a filter function that returns the detections with higher class confidences than $T_\text{conf}$. Here, $\mathbb{T}$ denotes the set of these thresholds, \textit{i.e.}, $\mathbb{T}=\{T_\text{IOU},T_\text{conf}\}$. Therefore, SMR for object detection represents the proportion of machines with higher mAPs than a reasonable threshold. 

We do not use human annotations when calculating satisfaction scores. Instead, the perception on the original image is used as ground truth, given that the perceptual differences between original and compressed images inherently expose MVS characteristics \cite{zhang2021just,jin2021just,fischer2022evaluation}. This approach allows for the study of SMR using a potentially unlimited number of images and videos in an unsupervised or self-supervised manner. However, this scheme may introduce some performance gaps when evaluating the machine's capability and coding efficiency in VCM \cite{fischer2022evaluation}. This issue is pronounced in object detection, where machines tend to generate excessive, often unreliable detections (many with low confidence scores) due to the development nature of them. Utilizing all detections from the original image without filtration leads to these redundant and unreliable detections becoming rigid benchmarks, which significantly interferes the complex mAP calculation of detections on the compressed image. To mitigate this, we introduce a customizable $\mathcal{F}_{T_\text{conf}}$ to selectively exclude inappropriate detections on the original image during satisfaction score calculation.

By aggregating satisfaction scores of many machines, SMR mitigates the perception biases from a few specific machines and measures the compression quality accurately with general MVS characteristics considered. Images with higher SMRs are more conducive to machine vision applications, as they are more likely to yield correct analysis results for most machines. The significance of SMR is further highlighted when the machine is unknown before compression, or in the long term the machine can be upgraded or replaced, which are common in the real world. More importantly, SMR can guide the compression to remove perceptual redundancy for MVS and preserve consistent machine perceptions on original and compressed images for most machines, achieving an ideal equilibrium between compression efficiency, generalizability, and machine analysis performance. Therefore, SMR is a more appropriate optimization target for VCM and fuels the perceptual coding for MVS.

\section{SMR Dataset} \label{sec:smr dataset}

In this section, we build the first SMR dataset to facilitate SMR studies. There are three steps: image preparation, machine library construction, and SMR annotation.

\subsection{Image preparation} \label{subsec:image preparation}

To construct a good SMR dataset, a large collection of high-quality images with different semantics are essential. Since these images are used to calculate satisfaction scores based on machine perception differences, it is reasonable to leverage existing, well-established machine vision task datasets. For this purpose, we select the MS COCO 2017 dataset. Similar to Section \ref{sec:machine diversity}, we compress all COCO images as intra video frames using HM-16.24. To cover a wide range of quality levels, we select 36 QPs, which are $11, 13, 15, 17, 19, 21, 22, \dots, \text{and } 51$.

\subsection{Machine library construction} \label{subsec:machine library construction}

To derive a practically applicable and precise SMR, we need to build a representative machine library $\mathbb{M}$ for the satisfaction score calculation and aggregation. There are several factors to consider for building $\mathbb{M}$, such as machine's capability, architecture, size, complexity, and the number of machines included. The machine's capability is the most important attribute that directly determines the task accuracy. However, it's important not to limit the library to only the most performant machines, as many less capable machines offer unique advantages and suit specific application scenarios. For instance, some machines have simpler structures that are easier to implement and accelerate with hardware. Some machines with much fewer parameters are more appropriate in stringent memory and computation complexity environments. Some other machines can run fast in low latency, satisfying real-time processing requirements. Moreover, many machines have already been widely used due to historical reasons and are costly to replace. Lastly, the number of machines should be sufficient to mitigate the perceptual bias from a few machines. In summary, as demonstrated in Section \ref{sec:machine diversity}, machine diversity should be meticulously factored into the library's construction to more comprehensively and accurately capture general MVS characteristics.

In this work, we conscientiously review the evolution and current landspace of the machine vision community and select several representative machines to construct machine libraries. For image classification, we initially collect 58 machines to build the v1 machine library. Then, to explore the impact of different numbers of machines on SMR distribution and prediction (which will be discussed later in Section \ref{subsec:dataset study} and Section \ref{subsec: generalize to more machines} to \ref{subsec: generalize to other datasets}), we further append 14 more machines to build the v2 library, containing 72 machines in total. Detailed information about these machines is presented in TABLE~I in the Appendix. The selected machines display a wide range of top-1 prediction accuracies on ImageNet (from $60\%$ to $88\%$), sizes (from $1.4$ to $300$ million parameters), and complexities (from $0.1$ to $360$ GFLOPs). Both cutting-edge transformers and aged CNNs, complex and lightweight machines, and hand-crafted and automatically-searched machines are included. We also consider machines in the same architecture but have different depths or scales. All machines are initialized with weights from PyTorch \cite{paszke2019pytorch} and fine-tuned on the original COCO train2017 dataset using cropped object images to align with COCO semantics.

For object detection, we build a machine library with 98 machines to embrace rapid developments in this field. Detailed information of these machines is presented in TABLE~II in the Appendix. Their mAPs on COCO val2017 dataset range from $21.3$ to $55.1$, with sizes ranging from $1.8$ to $200$ millions of parameters, and inference speeds from less than one frame-per-second (fps) to 1000 fps. Both CNN-based and transformer-based, one-stage and two-stage, performance-targeted and speed-targeted machines are included. We also include variations where the same detectors are equipped with different backbone networks. All machines are initialized with weights from MMDetection \cite{chen2019mmdetection} trained on the original COCO train2017 dataset.

Despite the rich diversity, the number of machines in the constructed libraries are notably larger compared to most HVS-oriented JND and SUR datasets, such as VideoSet \cite{wang2017videoset} (around $30$ human subjects per video) or KonJND-1k \cite{lin2022large} ($42$ subjects per image). We believe that it is adequate to study SMR with this library from a practical perspective. Note that considering fewer machines could bring several negative impacts, which will be described in Section \ref{subsec:dataset study}.

\subsection{SMR annotation} \label{subsec:smr annotation}

\begin{figure*}[htbp]
  \centering
  \subfloat{
      \includegraphics[width=0.46\textwidth]{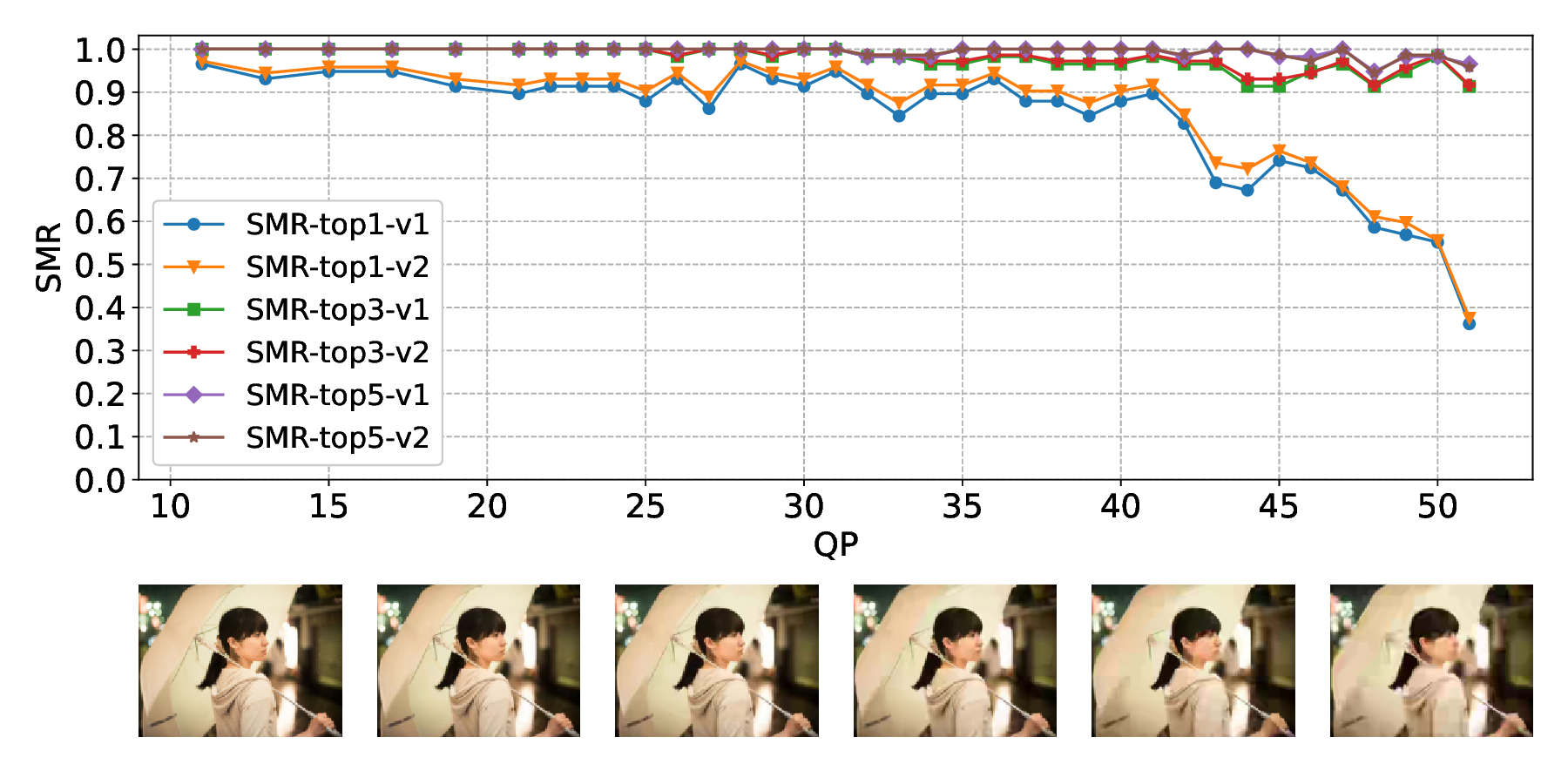}
  }
  \subfloat{
      \includegraphics[width=0.46\textwidth]{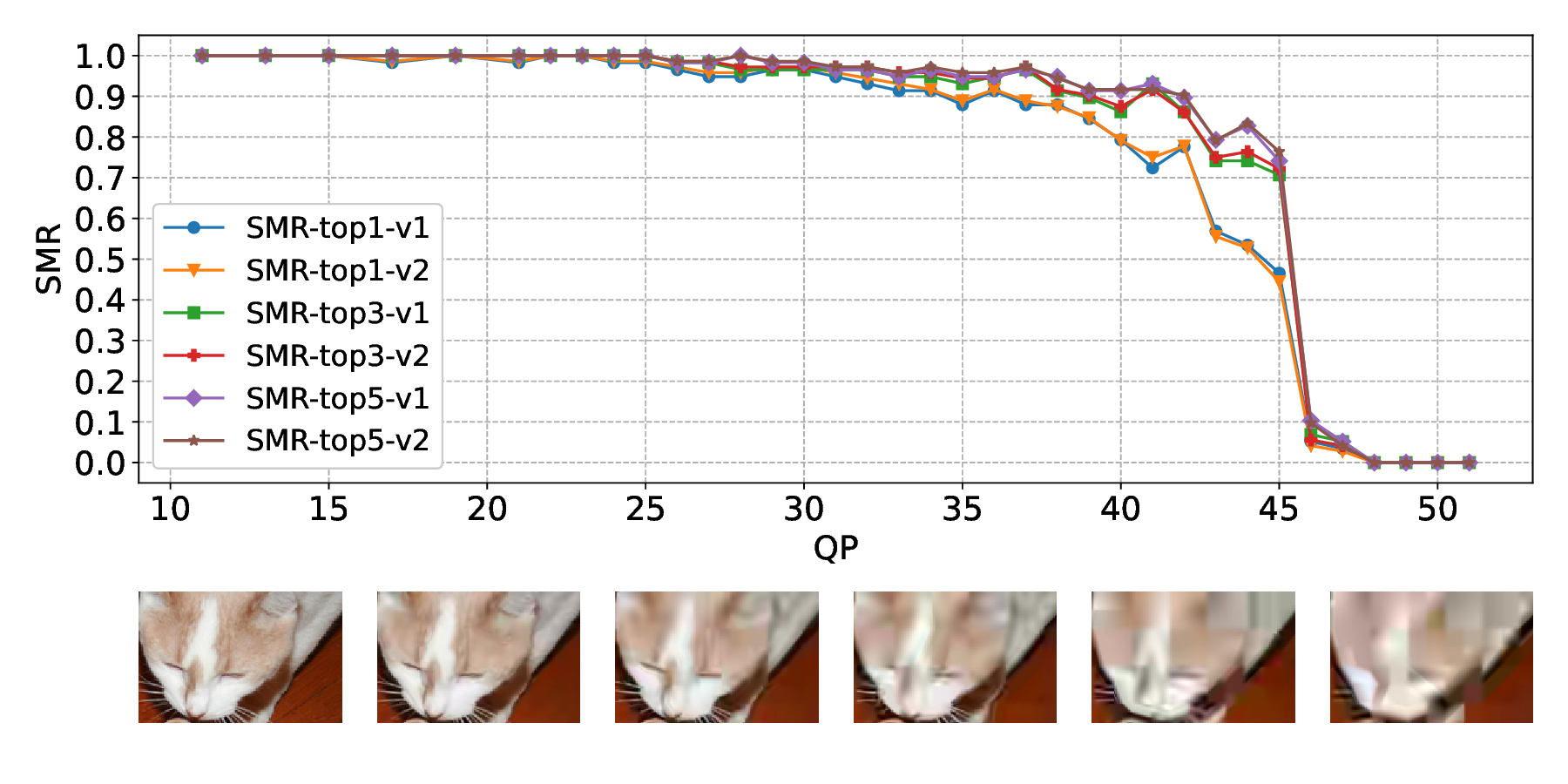}
  }
  \qquad
  \subfloat{
      \includegraphics[width=0.46\textwidth]{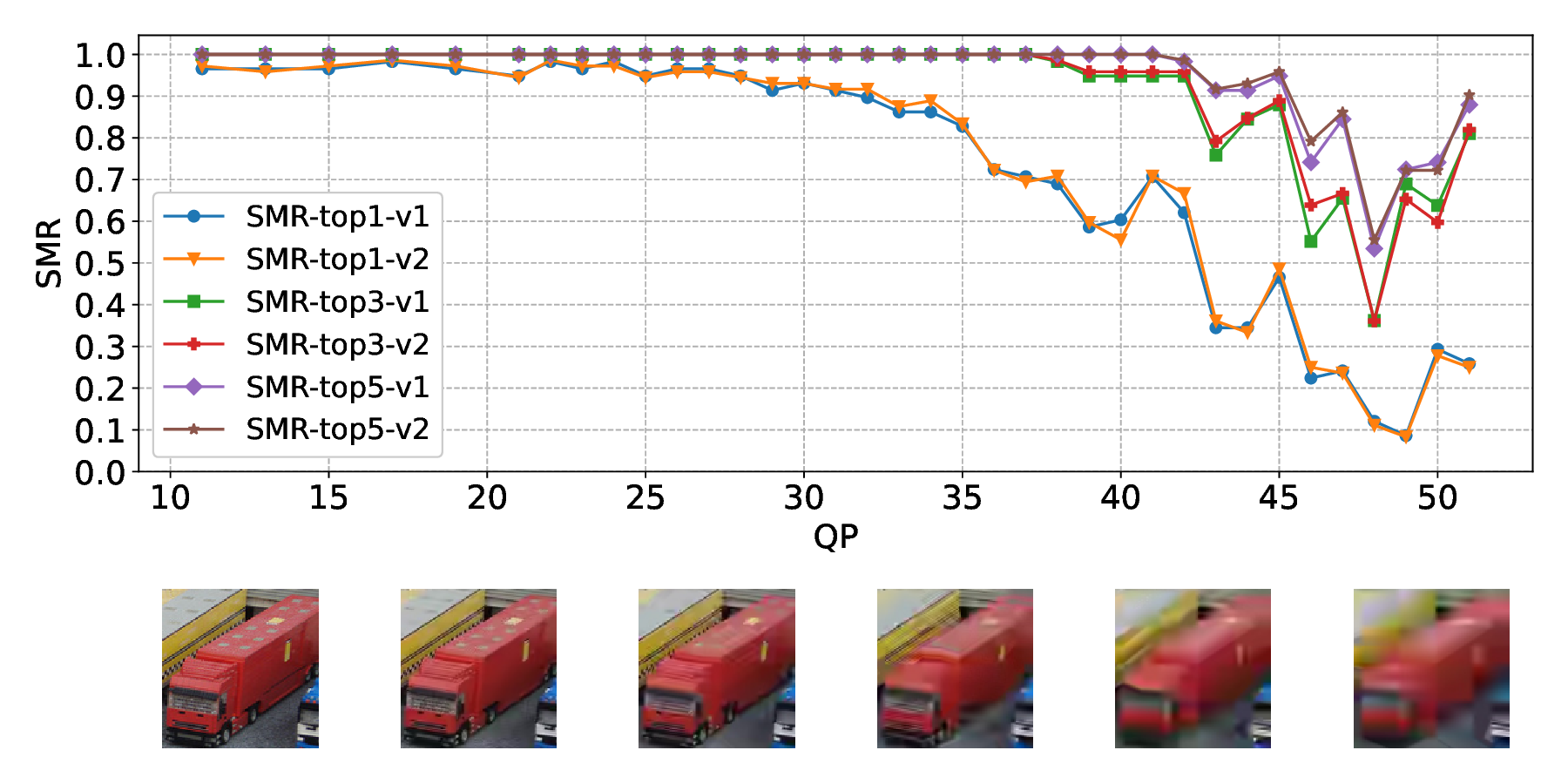}
  }
  \subfloat{
      \includegraphics[width=0.46\textwidth]{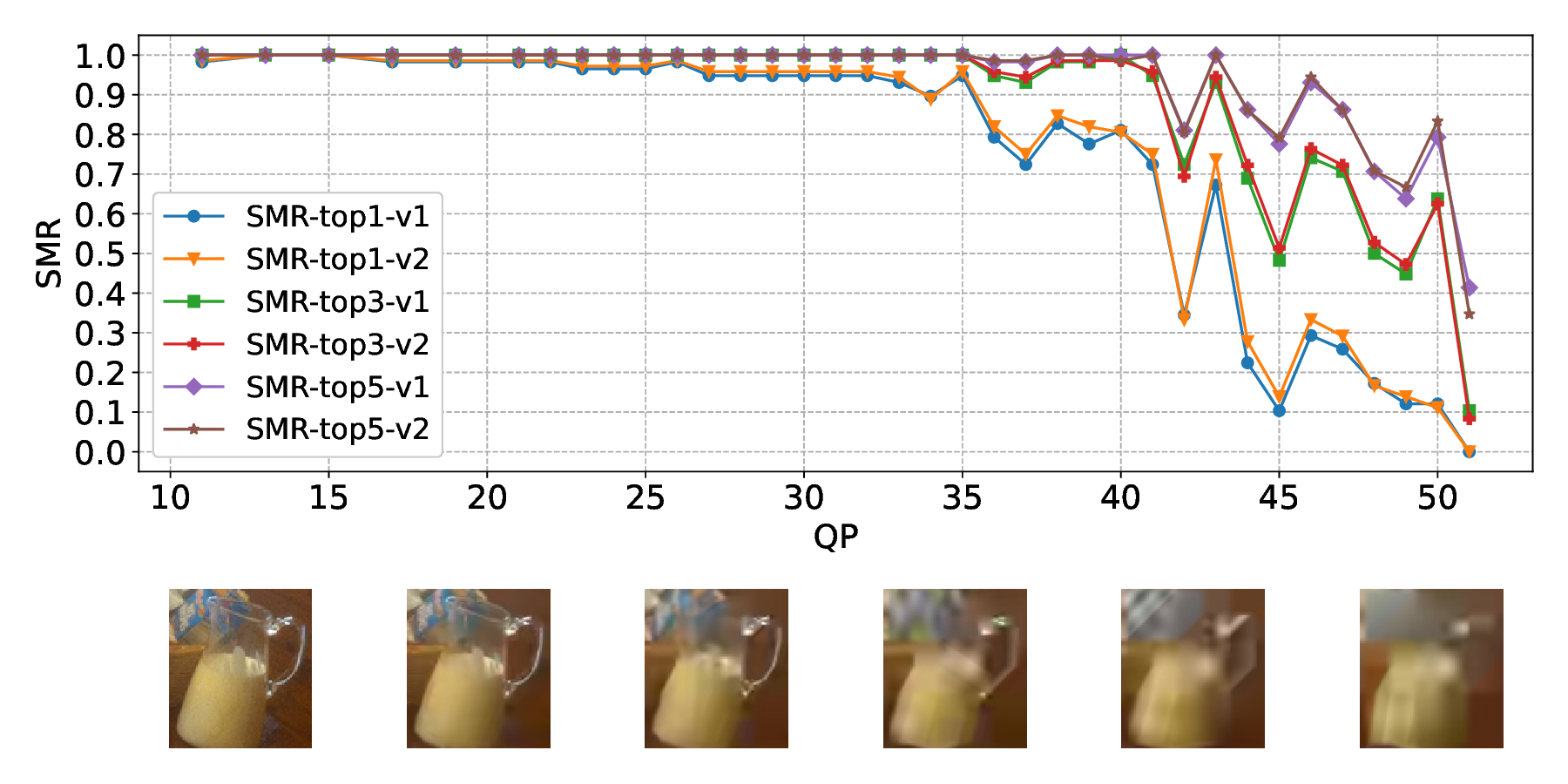}
  }
  \quad
  \subfloat{
      \includegraphics[width=0.46\textwidth]{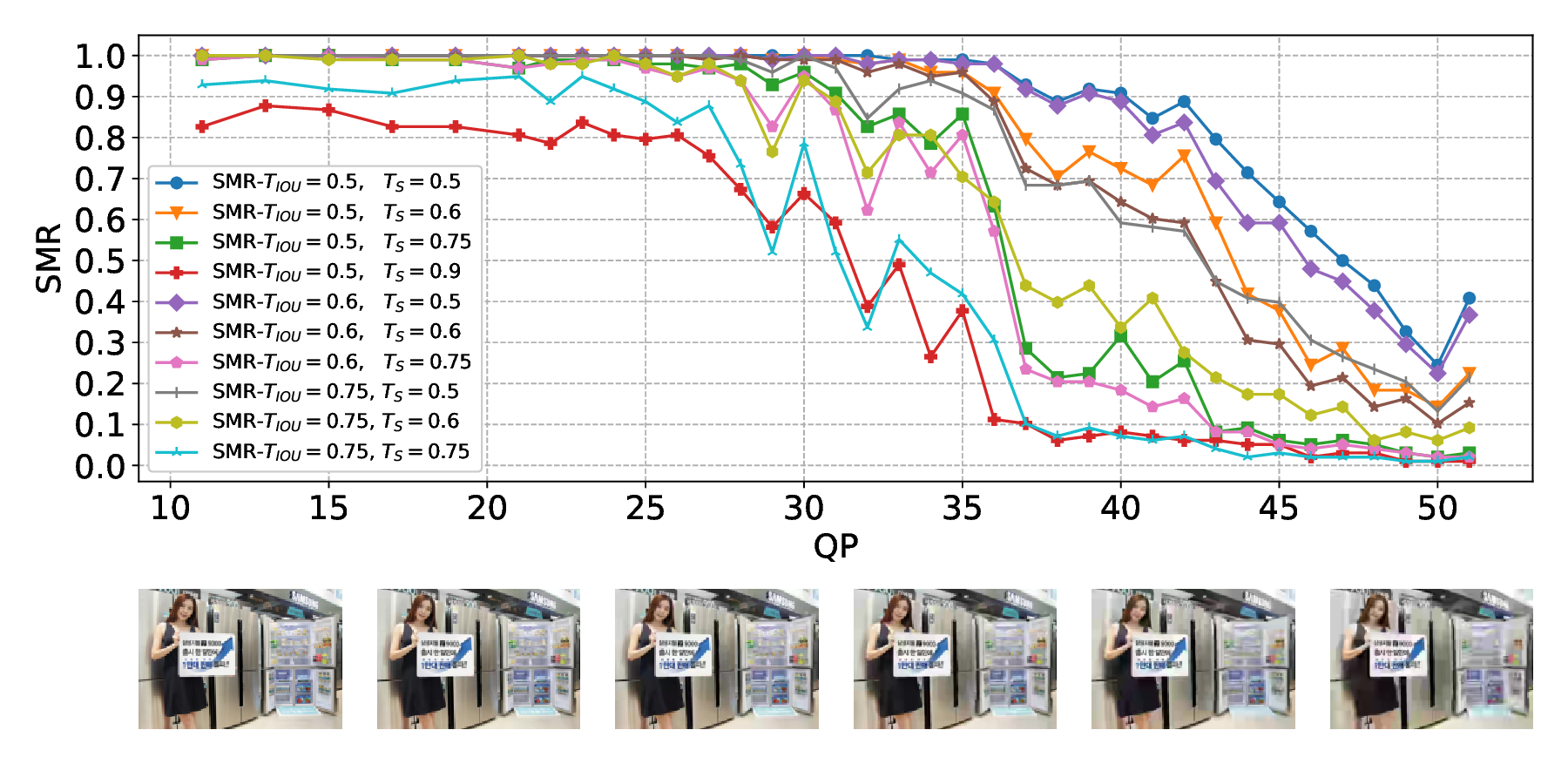}
  }
  \subfloat{
      \includegraphics[width=0.46\textwidth]{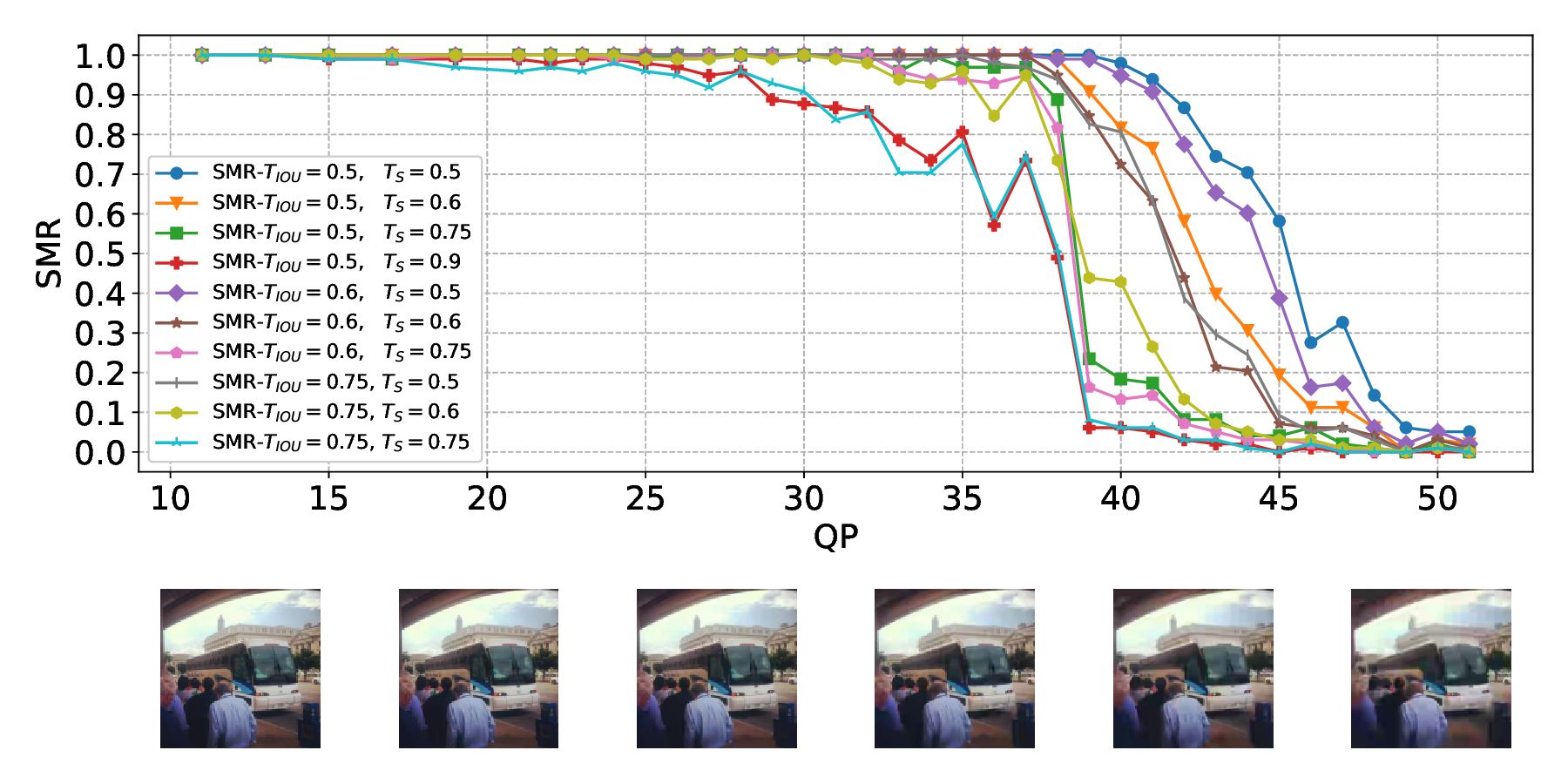}
  }
  \qquad
  \subfloat{
      \includegraphics[width=0.46\textwidth]{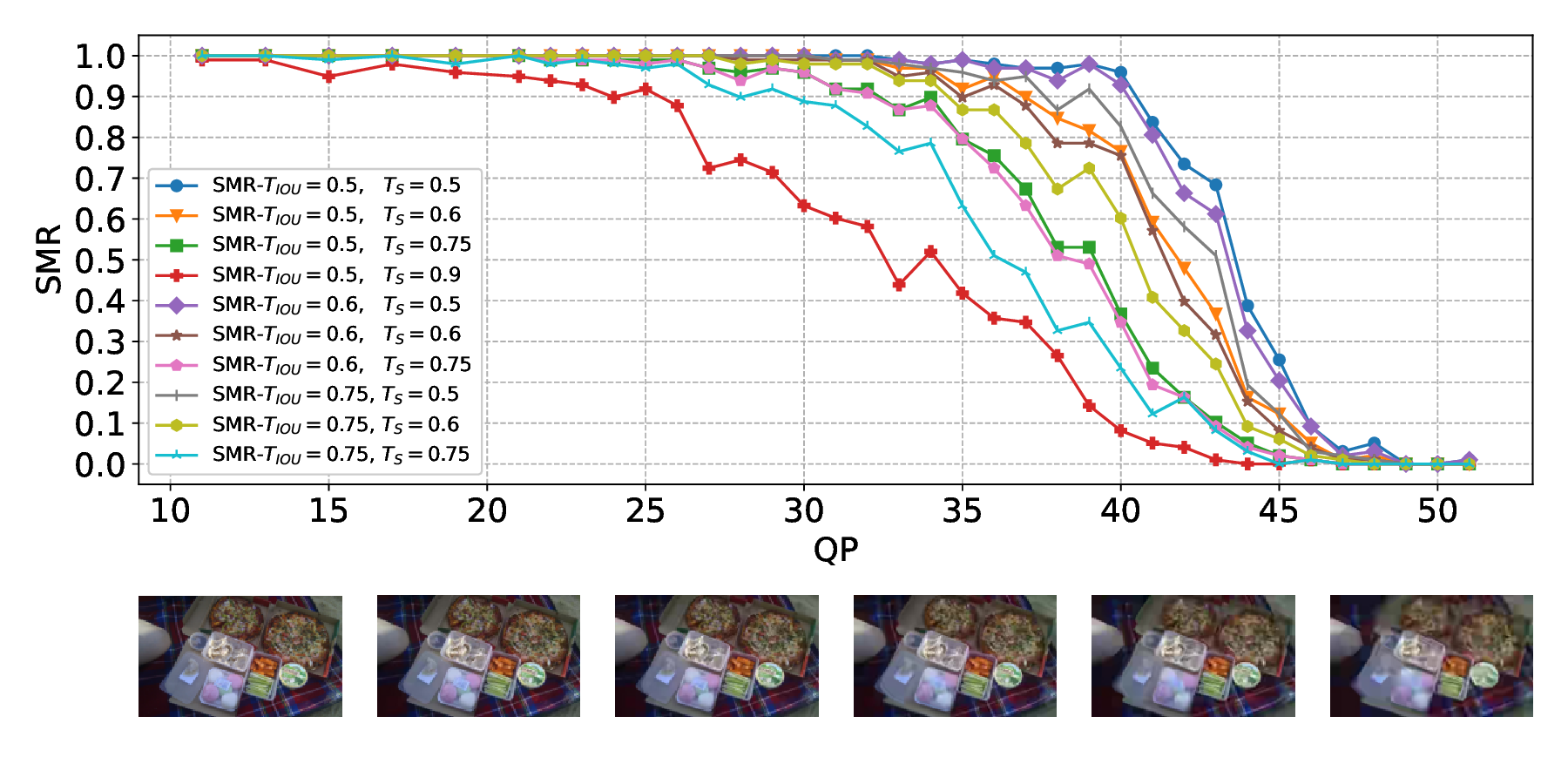}
  }
  \subfloat{
      \includegraphics[width=0.46\textwidth]{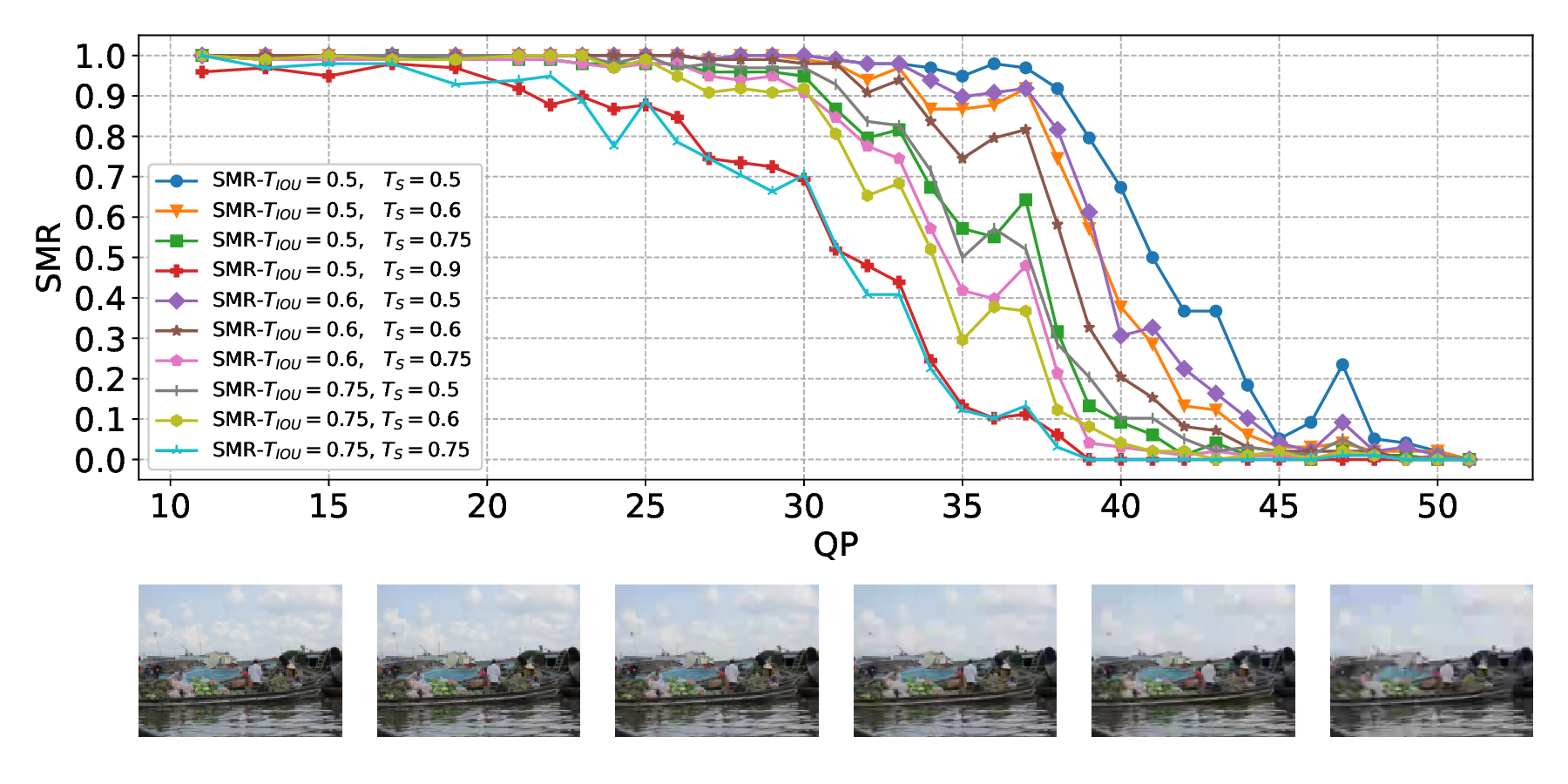}
  }
  
  \caption{QP-SMR curves of several randomly selected images. The corresponding image (or object image) and its compressed versions are displayed under the curve (better viewed by zooming in), where the leftmost one is the original object image, and the rest are compressed variants with $\text{QP}=32, 40, 45, 48, \text{and } 51$, respectively. The top 4 sub-figures are for image classification, and the bottom 4 are for object detection.}
  \label{fig:smr sample cls and det}
\end{figure*}

Due to the absence of annotations on the COCO test dataset, we randomly select $10,000$ original images from the COCO train2017 dataset to form our SMR test dataset, while retaining the existing validation set. After compression, each original COCO image or object image yields 36 distorted versions. We apply Eq. \eqref{eq:smr} to \eqref{eq:satisfaction score function for det} to generate ground truth SMR labels for all images, including the originals. For image classification, SMR labels are annotated for object images larger than $32\times32$ pixels. We set $K=1,3,5$ in Eq. \eqref{eq:satisfaction score function for cls} and use two versions of machine libraries for the annotation separately, resulting in six types of SMR, namely SMR-top1/3/5-v1/v2. For object detection, SMR labels are annotated for full images. We set $T_\text{IOU}$ to $[0.5:0.05:0.95]$ (which means values range from $0.5$ to $0.95$ with a step of $0.05$) and $T_\text{conf}$ to $0.3$ in Eq. \eqref{eq:satisfaction score function for det}, and set $T_S$ to $[0.5:0.05:0.95]$ in Eq. \eqref{eq:smr}, resulting in $10\times10=100$ types of SMR. Eventually, for image classification, the constructed SMR dataset contains $617,479\times(36+1)=22,846,723$ images and more than $137$ million SMR labels. There are $20,052,335$, $936,396$, and $1,857,992$ images for train, validation, and test dataset, respectively. For object detection, the constructed SMR dataset contains $123,287\times(36+1)=4,561,619$ images and more than $456$ million SMR labels. There are $4,006,619$, $185,000$, and $370,000$ images for train, validation, and test dataset, respectively. The scale of our SMR dataset is marvelous. We are confident it will establish a solid foundation for both this work and future SMR research.

\subsection{Dataset study} \label{subsec:dataset study}

\begin{figure}[htbp]
  \centering
  \subfloat[Image classification]{
      \includegraphics[width=0.48\textwidth]{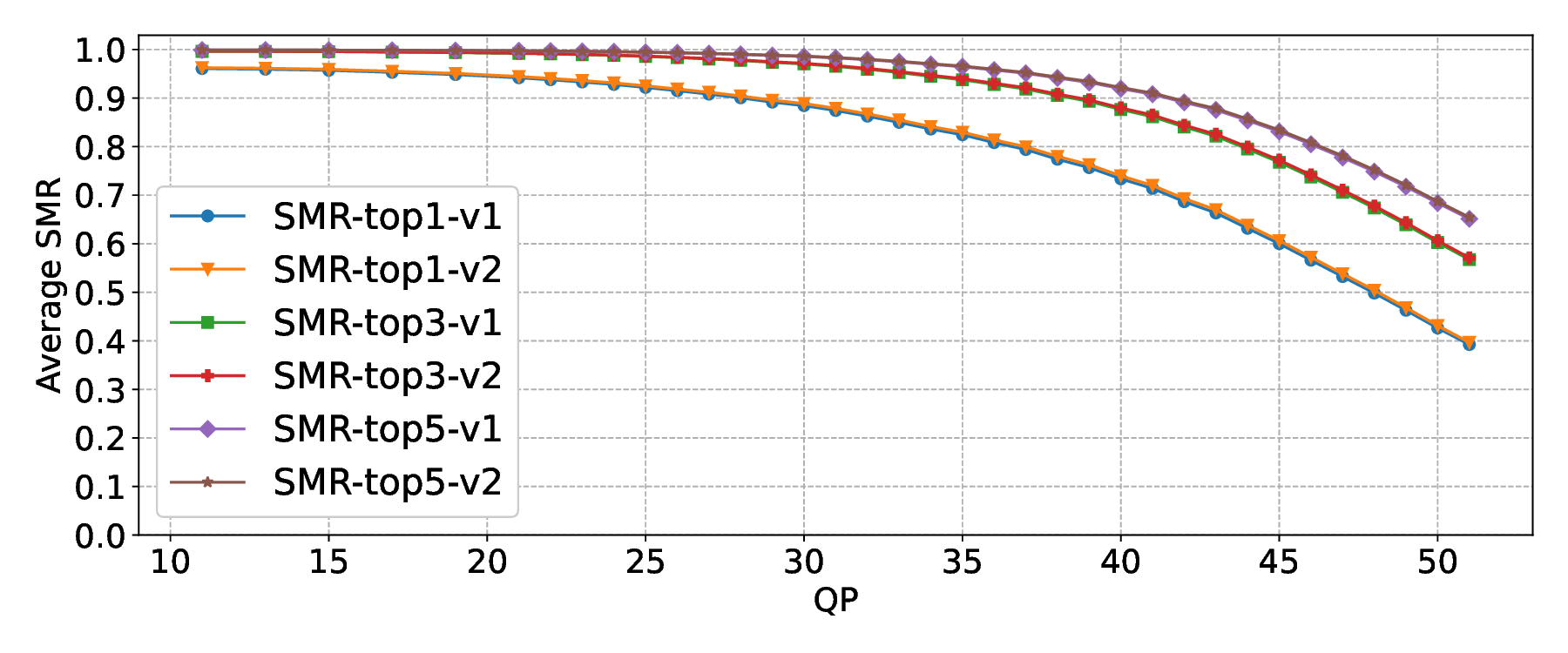}
  }
  \quad
  \subfloat[Object detection]{
      \includegraphics[width=0.48\textwidth]{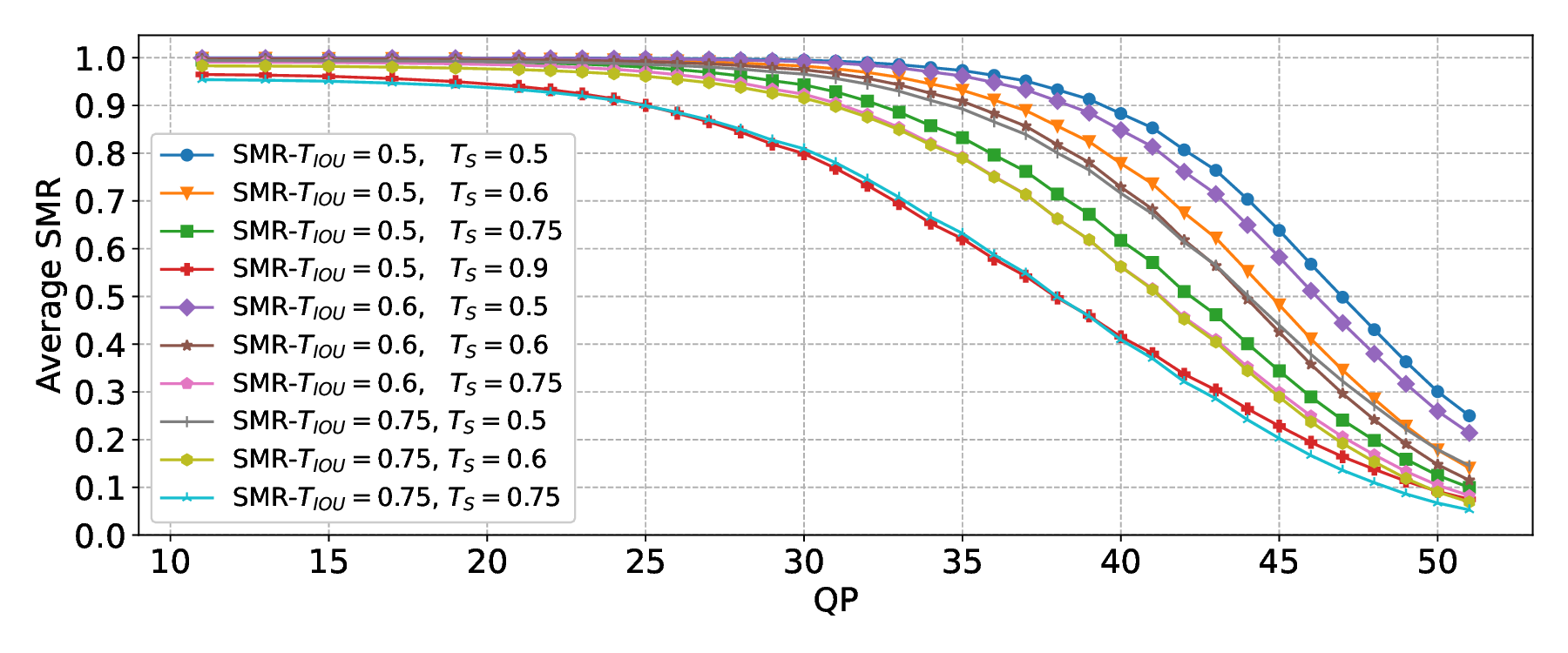}
  }
  \caption{QP-SMR distributions of the SMR dataset for different tasks.}
  \label{fig:smr distribution}
\end{figure}

We plot several QP-SMR curves of randomly selected images in Fig.~\ref{fig:smr sample cls and det} to give a first impression of how SMR curves look. It is obvious that different images have different QP-SMR curves, thus the image's SMR is a content-related attribute. The SMR distribution of our dataset, \textit{i.e.} average SMR values under different QPs, are presented in Fig.~\ref{fig:smr distribution}. Both the individual QP-SMR curves and the overall SMR distributions underscore the prevalence of machine diversity again because if machines had identical MVS characteristics, SMR would consistently be either $0$ (no machine is satisfied with the compression quality) or $1$ (every machine is satisfied), which is opposite against the truth.

Generally, SMR tends to decrease as QP increases, yet the amplitude changes across different QPs. For lower QPs, there is typically a slight decrease in SMR, while at higher QPs, SMR often drops noticeably and becomes unusable. Interestingly, there can be significant SMR variations even between adjacent QPs, indicating that QP and SMR do not have a monotonic correlation. In other words, a higher QP may sometimes lead to a higher SMR, implying that lower compression quality could result in more consistent machine perceptions. This observation holds substantial importance for two reasons. First, \textbf{many machines exhibit similar perceptual instability when facing compression distortions}. It is expected that a single specific machine might have its flaw that cannot precisely distinguish good and bad compression quality. However, this uncertainty is anticipated to be erased when many machines are considered together, which is surprisingly not true. Second, it reveals that \textbf{achieving higher SMR under heavier compression is feasible}, which is ideal for VCM as both lower bit-rate and higher analysis performance for many machines can be achieved simultaneously.

The characteristics and distributions of different types of SMR are distinct. For image classification, for a given QP, $\text{SMR-top5} \geq \text{SMR-top3} \geq \text{SMR-top1}$. For object detection, SMR is jointly controlled by $T_\text{IOU}$ and $T_S$, with $T_S$ typically exerting a greater impact. Moreover, the trends in SMR variations also differ across SMR types. For instance, $\text{SMR-top5}(I_{q_i})>\text{SMR-top5}(I_{q_j})$ does not necessarily lead to $\text{SMR-top1}(I_{q_i})>\text{SMR-top1}(I_{q_j})$.

\bgroup
\def\arraystretch{1.2}
\begin{table}[htbp]
  \centering
  \caption{Differences between SMRs obtained from various number ($N_m$) of machine subjects and from v1/v2 machine library.}
  \label{table:smr mae of different amount of machines}
  \resizebox{0.48\textwidth}{!}{
  \begin{tabular}{c|cccccccc}
  \Xhline{4\arrayrulewidth}
  $\mathbf{N_m}$ & \textbf{10} & $\textbf{12}^*$ & \textbf{20} & \textbf{30} & \textbf{40} & \textbf{50} & \textbf{60} & \textbf{70} \\
  \Xhline{3\arrayrulewidth}
  \textbf{58 (v1)} & 0.051 & 0.047 & 0.034 & 0.028 & 0.022 & 0.016 & 0.013 & 0.012 \\ \hline
  \textbf{72 (v2)} & 0.051 & 0.044 & 0.033 & 0.025 & 0.019 & 0.013 & 0.009 & 0.004 \\
  \Xhline{4\arrayrulewidth}
  \end{tabular}
  } 
\end{table}
\egroup

In comparing the v1 and v2 machine libraries for image classification, we find their SMR values relatively similar. Numerically, within the SMR dataset (excluding uncompressed images), the mean absolute errors (MAE) for SMR-top1/3/5 between v1 and v2 libraries are $0.011$, $0.007$, and $0.005$, respectively, with standard deviations of $0.024$, $0.016$, and $0.014$. Considering specific images, it can also be observed from Fig.~\ref{fig:smr sample cls and det}-\ref{fig:smr distribution} that SMR values are quite close between the two libraries. Despite minor numerical differences, the general trends in SMR variation remain largely consistent. We further conduct a comparative analysis of SMRs derived from different numbers of machines. In this study, machines are randomly picked from the v2 library to re-calculate SMR-top1 values, denoted as $\text{SMR}_{N_m}$, where $N_m$ represents the number of machines. We also investigate a special case using the 12 machines in Section \ref{sec:machine diversity}. The MAEs of $\text{SMR}_{N_m}$ values are computed among $10,000$ groups of compressed images that are also randomly selected, each in all 36 quality levels. For each $N_m$, the comparison will be performed three times to obtain averaged results, which are presented in TABLE~\ref{table:smr mae of different amount of machines}. The results indicate that $\text{MAE}(\text{SMR}_{N_m}, \text{SMR}_{58/72})$ decreases as $N_m$ increases, although minor differences persist. Smaller $N_m$ values inevitably lead to less accurate general MVS characteristics capturing due to machine diversity. Again, this demonstrates that we cannot just consider one or only very few of them in VCM.

\subsection{SMR vs. JND} \label{subsec:smr vs jnd}

During SMR calculations, satisfaction score functions determine whether a machine is satisfied with compression quality based on the perceptual differences between original and compressed images. These functions appear to be capable of locating JND for MVS in terms of QP: for a machine $M_m$ and an original image $I_i$, JND is expected to locate at the minimum QP $q_\text{min}$ where $\text{S}(M_m;I_{q_\text{min}})<T_S$. Meanwhile, for any higher QP $q_x>q_\text{min}$, $\text{S}(M_m;I_{q_x})<T_S$ should consistently hold. However, we find that for many machines on many images, it is not true.

\bgroup
\def\arraystretch{1.2}
\begin{table}[htbp]
  \centering
  \caption{Proportion of images in percentage where at least $k_1$ machines have more than $k_2$ PC variations.}
  \label{table:k1 machine k2 perception variation}
  \resizebox{0.48\textwidth}{!}{
  \begin{tabular}{l|cccccc}
  \Xhline{4\arrayrulewidth}
  & \textbf{2} & \textbf{3} & \textbf{4} & \textbf{5} & \textbf{6} & \textbf{7} \\
  \Xhline{3\arrayrulewidth}
  \textbf{1} & 95.89/85.86 & 92.85/77.71 & 89.72/70.81 & 86.44/63.76 & 80.82/53.58 & 77.44/47.43 \\ \hline
  \textbf{3 (5\%)} & 91.47/76.71 & 86.69/66.25 & 81.16/56.70 & 76.44/48.26 & 67.31/36.59 & 63.03/30.77 \\ \hline
  \textbf{6 (10\%)} & 88.32/70.29 & 82.32/58.48 & 74.53/46.87 & 69.25/39.16 & 57.59/26.58 & 53.35/21.55 \\ \hline
  \textbf{18 (30\%)} & 79.31/54.10 & 71.53/42.00 & 56.54/26.24 & 51.08/20.21 & 35.27/9.76 & 31.23/6.89 \\ \hline
  \textbf{29 (50\%)} & 69.92/41.07 & 61.83/29.95 & 41.18/14.02 & 35.91/9.57 & 18.94/3.07 & 14.88/1.59 \\ \hline
  \textbf{41 (70\%)} & 55.09/24.82 & 46.62/16.29 & 21.40/4.22 & 16.48/2.12 & 4.97/0.28 & 2.95/0.10 \\ \hline
  \textbf{53 (90\%)} & 22.40/4.81 & 14.49/2.03 & 2.42/0.15 & 1.04/0.04 & 0.12/0.00 & 0.03/0.00 \\
  \Xhline{4\arrayrulewidth}
  \end{tabular}
  } 
\end{table}
\egroup

To investigate the JND characteristics of MVS more comprehensively, we utilize the \textit{\textbf{machine perception consistency sequence}} in Section \ref{sec:machine diversity}. A key indicator of JND presence for a machine is the occurrence of exactly one ``10'' and no ``01'' in its Perception Consistency (PC) sequence. We record such sequences of all 58 machines in the v1 machine library on the SMR validation dataset, counting the occurrences of ``10'' and ``01''. We compute PC based on top1/5 prediction like Eq. \eqref{eq:satisfaction score function for cls}, utilizing all 36 QPs. Our analysis reveals that the numbers of ``10'' and ``01'' (called PC variations and denoted by $N_\text{var}$) averaged among all machines and images are 3.68/1.81 for top1/5-based PC calculation, respectively. We further assess the proportion of images where $N_\text{var} \geq k_2$ for at least $k_1$ machines. The results are shown in TABLE~\ref{table:k1 machine k2 perception variation}, where the first column lists $k_1$ values and corresponding machine proportions, and the first row lists $k_2$ values. Each table cell shows two proportion values for top1/5-based PC calculation, respectively. On most images (about $96\%/86\%$), at least one machine has 2 or more PC variations, suggesting the absence of JND. On around $70\%/41\%$ of images, $N_\text{var} \geq 2$ is satisfied for $50\%$ of machines. Therefore, it is common for machines not to have JND on many images when image quality deteriorates due to compression. Meanwhile, on over $51\%/20\%$ of images, $30\%$ of machines have 5 or more PC variations, indicating that frequent PC variations across different compression quality levels occur commonly among various machines.

We further focus on two special cases where perceptions on all 36 compressed images of at least one machine are 1) uniformly consistent and 2) entirely divergent from the perception on the original image. From the results, $63.51\%/19.18\%$ of images fall into case 1/2 under top1-based PC calculation, and $93.65\%/0.44\%$ under top5-based PC calculation, respectively. Furthermore, on $2.79\%/10.77\%$ of images, all machines satisfy case 1 under top1/5-based PC calculation. However, no image can let all machines satisfy case 2. This observation is intriguing, particularly when compared to JND for HVS, as HVS can always overlook minor quality differences at lower QPs and perceive significant ones at higher QPs. Therefore, on specific images, several machines demonstrate either a heightened resilience or vulnerability to compression quality degradations, both leading to JND's absence.

Note that HVS can have multiple JNDs \cite{jin2016statistical,wang2017videoset,tian2020just}, but only the first JND is determined by using the original image as reference, and subsequent JNDs are referencing the preceding JND image. In contrast, we reveal that MVS can have multiple PC variations by consistently referencing the original image. Moreover, our conclusions do not contradict pioneering works \cite{zhang2021just,jin2021just,zhang2023learning} exploring JND for MVS. The reason is that they primarily consider the minimum noticeable distortion for MVS but ignore the possibility of tolerance to heavier distortions. As a comparison, we consider the entire QP range to offer a more comprehensive perspective. Importantly, the presence or absence of JND for MVS will not influence SMR, as SMR is defined independently of JND.

\subsection{SMR vs. SUR}

\begin{figure}[htbp]
  \centering
  \includegraphics[width=0.48\textwidth]{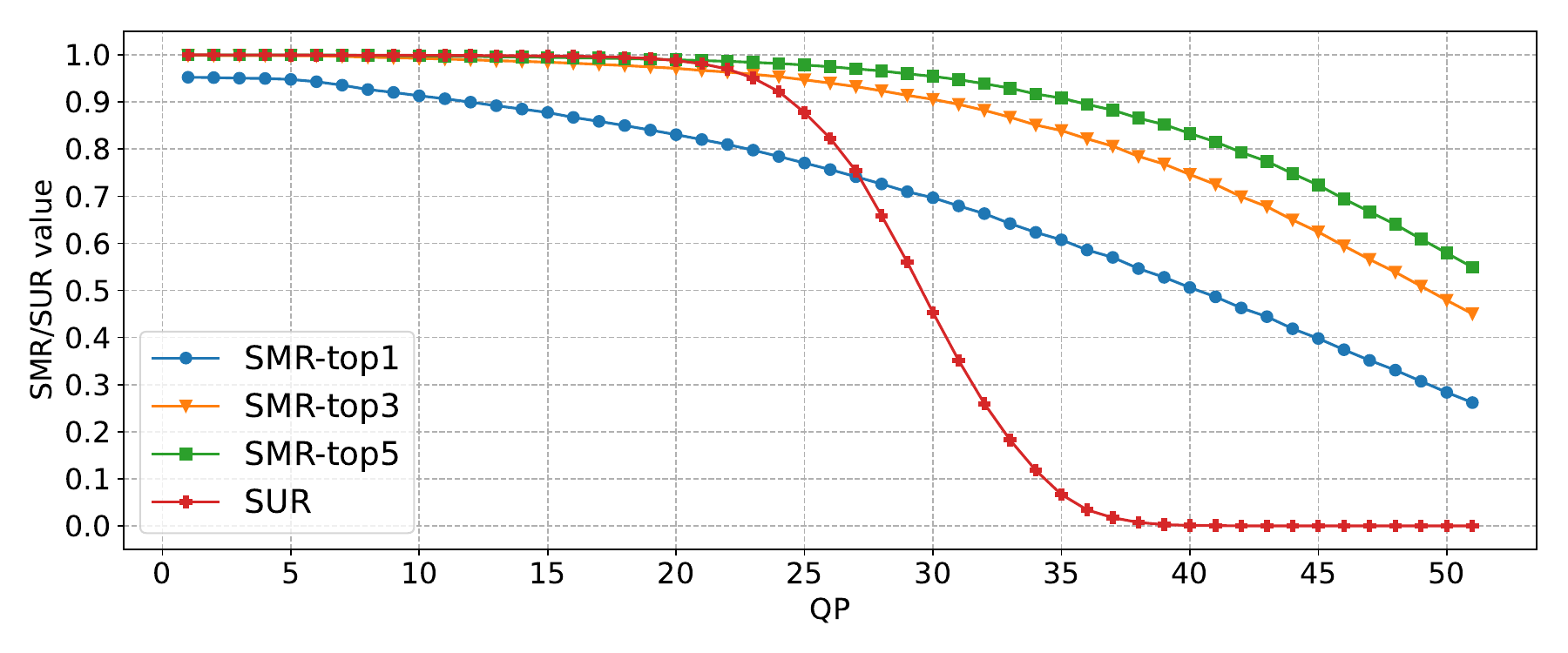}
  \caption{QP-SMR/SUR distributions of the KonJND-1k dataset \cite{lin2022large}.}
  \label{fig:qp-smr/sur konjnd}
\end{figure}

We make a simple experiment to elucidate the differences between SMR and SUR. Specifically, we annotate SMR for image classification on a large-scale JND/SUR dataset KonJND-1k \cite{lin2022large} using the v1 machine library. Due to the absence of object annotations on this dataset, we crop the original images into evenly sized patches of $224\times224$ pixels and annotate SMR for each patch. The resulting QP-SMR/SUR curves are depicted in Fig.~1, clearly indicating distinct distributions between SMR and SUR. Notably, SUR reaches 1.0 at very low QPs while not all types of SMR achieve this level, demonstrating the greater stability of HVS to tiny compression distortions. Conversely, at higher QPs, SUR drops to 0.0, while all types of SMR do not, highlighting a preferable robustness of MVS to heavy compression distortions. Furthermore, the decline in SUR is more rapid than that of SMR, indicating the HVS's heightened sensitivity to quality degradations, particularly at mid-range QPs. On the other hand, SMR exhibits a more gradual and steady decrease.

\section{SMR Modeling} \label{sec:smr modeling}

SMR is a valuable image quality metric for VCM to evaluate the machine perception consistency for the majority of machines under different compression quality levels. This evaluation helps improve the VCM encoder to remove perceptual redundancy for MVS by identifying the most efficient coding parameters, achieving higher SMR values under a similar bit-rate. However, in practical scenarios, collecting satisfaction scores from every machine in the library and accurately aggregating them into SMR is excessively time- and resource-intensive. Therefore, predicting the SMR of an image or video frame using a simpler model is crucial. In this work, we propose the task of SMR prediction or SMR modeling for VCM in the following formulation: with image $I_{q_i}$ in compression quality level $q_i$ being the distorted variant of the original image $I_0$, the objective is to develop a model $G_\theta$ such that

\begin{equation}
  G_\theta(\mathcal{I}) = \text{SMR}(I_{q_i}),
\end{equation}

\noindent where $\theta$ represents the model's learnable parameters, and $\mathcal{I}$ is the set of input images, which can either be $\mathcal{I}=\{I_{q_i}, I_0\}$ or $\mathcal{I}=\{I_{q_i}\}$. This formulation leads to two variants of the SMR modeling task: full-reference and no-reference SMR modeling, with the ``reference'' being the original image $I_0$.

In this work, we make an initial attempt on full-reference SMR modeling. Since SMR is a continuous decimal value within the range of $[0, 1]$, SMR modeling can be regarded as a regression task. Acknowledging SMR's content-specific nature, we propose a data-driven method based on deep learning to predict SMR independently for any image or video frame. Specifically, our SMR model comprises two components: an encoder $E(\cdot)$ and a regressor $R(\cdot)$. The encoder $E(\cdot)$ encodes the input image $x \in \mathbb{R}^{3 \times H \times W}$ into a latent representation $h \in \mathbb{R}^d$, where $H$ and $W$ are height and width of the image, and $d$ is the number of dimensions of $h$. The regressor $R(\cdot)$ then takes these representations from both original and compressed images as inputs, and outputs an estimated SMR of the compressed image.

\begin{figure}[htb]
  \centering
  \subfloat{
      \includegraphics[width=0.23\textwidth]{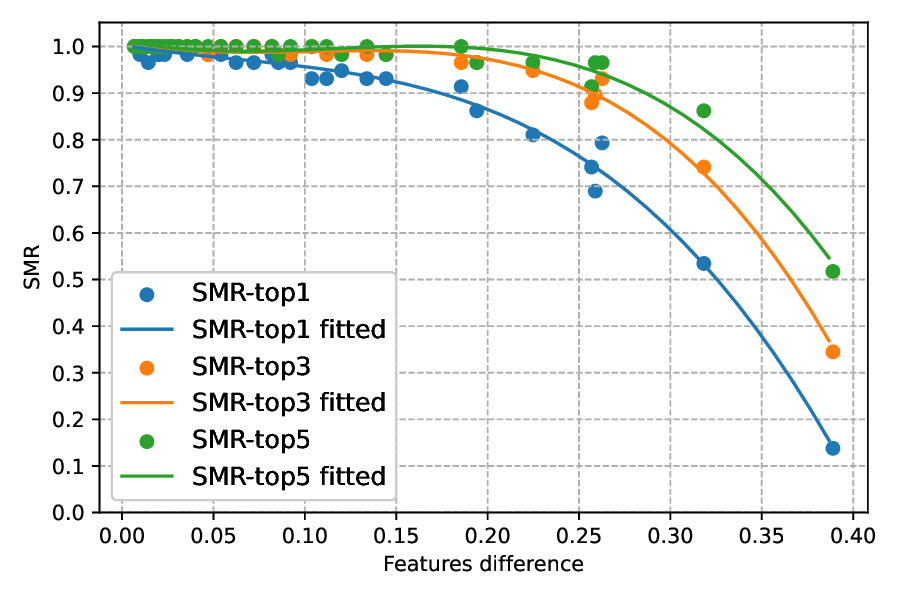}
  }
  \subfloat{
      \includegraphics[width=0.23\textwidth]{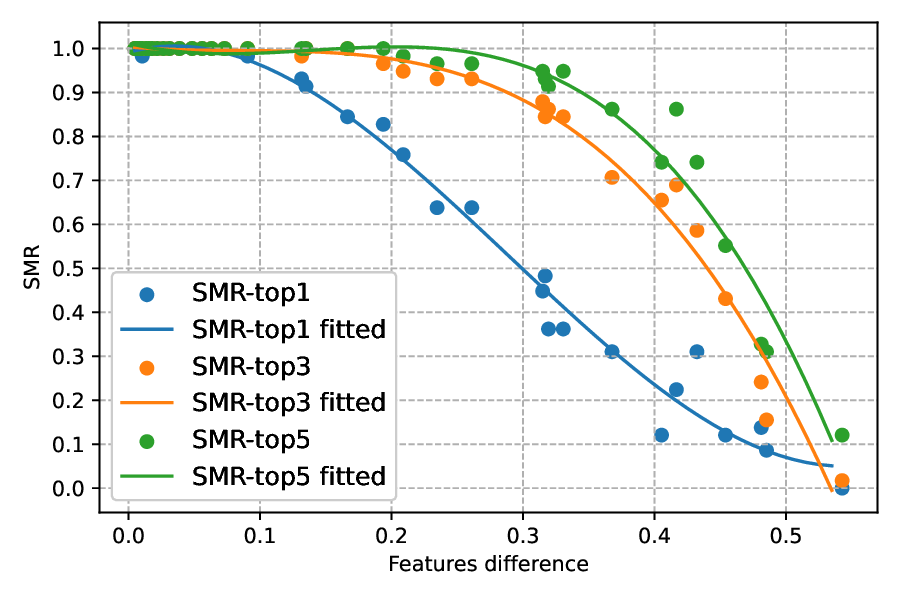}
  }
  \quad
  \subfloat{
      \includegraphics[width=0.23\textwidth]{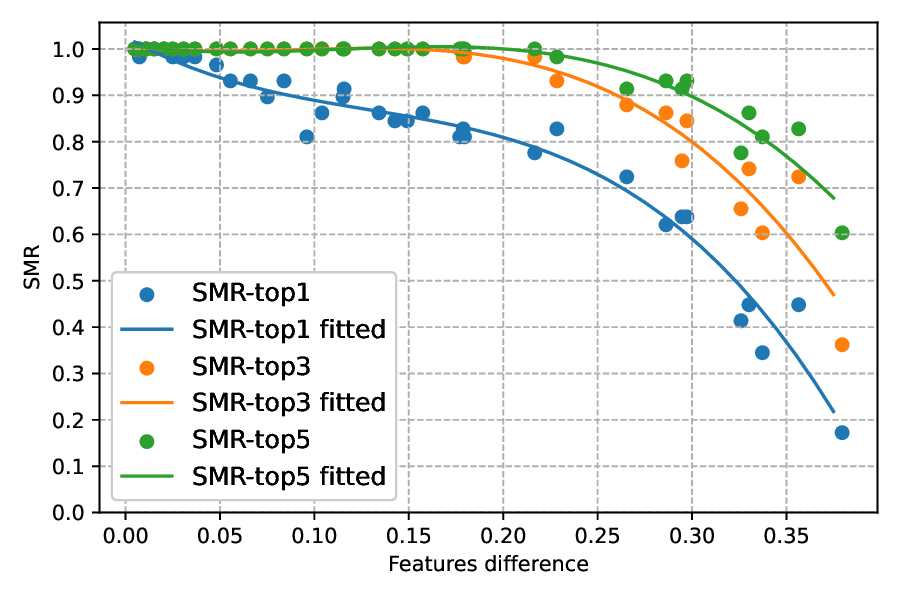}
  }
  \subfloat{
      \includegraphics[width=0.23\textwidth]{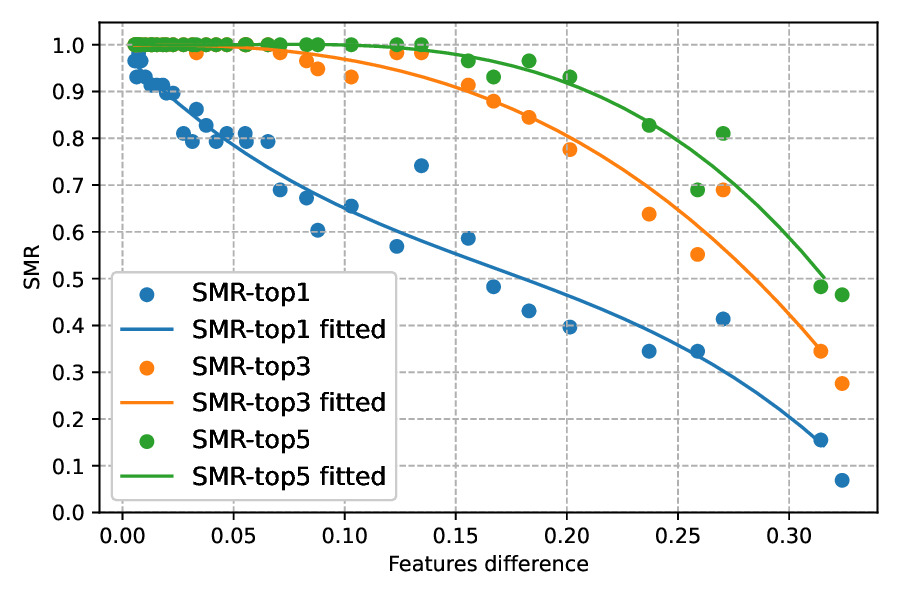}
  }
  \caption{The non-linear negative correlations between mean feature differences and SMR for several images in the SMR dataset.}
  \label{fig:features-diff-smr curves}
\end{figure}

The crux of our approach lies in learning a distinguishable and robust representation capable of predicting the similarity of machine perceptions across varying compression quality levels. Given that machines perceive and analyze images by extracting high-level semantic features, it is reasonable to hypothesize that deep features can inherently serve as good representations for this task and strongly correlate with SMR values. To test this hypothesis, we first extract deep features from image $I_0$ and all compressed variants using every machine in the v1 machine library. Then, for each compression quality level, we calculate the cosine distance between the original and compressed features as the feature difference, which will be averaged among all machines. After that, we examine the correlation between mean feature difference and SMR. As illustrated in several samples in Fig.~\ref{fig:features-diff-smr curves}, where the curves are fitted cubically as a reference, a consistent non-linear negative correlation between these two variables is verified. This observation validates our premise that deep features are suitable representations for SMR modeling.

\begin{figure}[htbp]
  \centering
  \includegraphics[width=0.48\textwidth]{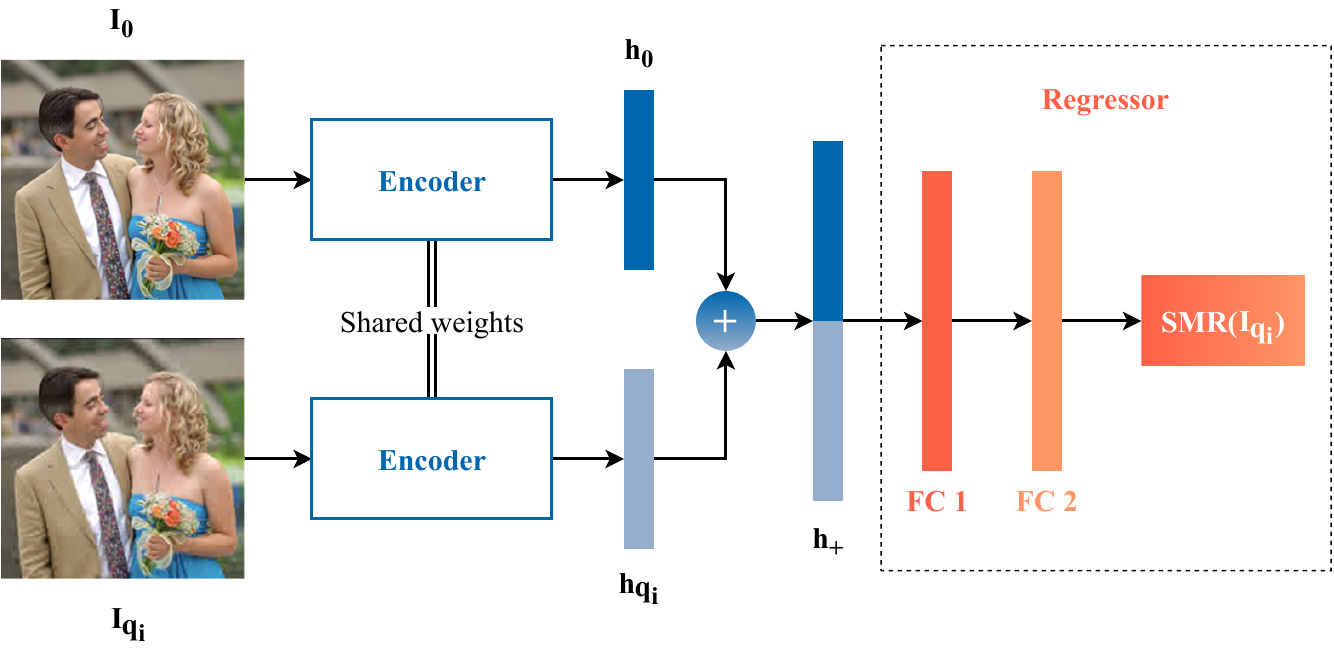}
  \caption{The proposed full-reference SMR prediction model.}
  \label{fig:smr model}
\end{figure}

Based on the investigation, we propose a straightforward yet effective baseline model $G$ for the SMR prediction task, depicted in Fig.~\ref{fig:smr model}. The model's encoder $E(\cdot)$ is a Siamese neural network trained for machine vision tasks like image classification but with fully-connected (FC) layers removed. This encoder extracts features from $I_0$ and $I_{q_i}$ to generate the corresponding latent representations $h_0$ and $h_{q_i}$. We then concatenate $h_0$ and $h_{q_i}$ to create an embedding $h_+ \in \mathbb{R}^{2d}$ that implicitly contains the feature difference information. Finally, this combined embedding $h_+$ is fed into the regressor $R(\cdot)$, which is designed as a multi-layer perceptron (MLP) network. The MLP is tasked with learning the non-linear relationship between feature differences and SMR. The training objective is to minimize the L1 distance between the predicted and actual SMR values:

\begin{equation}
  \min |R(E(I_0) \oplus E(I_{q_i})), \text{SMR}(I_{q_i})|.
\end{equation}

A significant limitation of the baseline SMR prediction model $G$ is its underutilization of available labeled data. Specifically, it only leverages the information from the SMR difference between the original image and its compressed version. However, we can actually further utilize the SMR difference between the two compression variants to learn better image representations for capturing the SMR characteristics. Hence, we design another model $Q_\phi$ such that

\begin{equation}
  Q_\phi(I_{q_i}, I_{q_j}) = |\text{SMR}(I_{q_i}) - \text{SMR}(I_{q_j})|,
\end{equation}

\noindent where $\phi$ is the model's learnable parameters. The architecture and training objective of $Q_\phi$ remain the same as the baseline model, as this auxiliary task is also a regression task accomplished by extracting good embeddings. After the training is finished, this SMR difference-based model $Q$ can predict the SMR of $I_{q_i}$ by:

\begin{equation}
  \hat{\text{SMR}}(I_{q_i}) = 1 - Q(I_0, I_{q_i}).
\end{equation}

\section{Experiments} \label{sec:experiments}

In this section, we conduct extensive experiments to validate the effectiveness and generalizability of our SMR models. We use \textit{cls} and \textit{det} as abbreviations for the image classification and object detection task, and model $G$ and $Q$ for the baseline and SMR difference-based model, respectively.

\subsection{Implementation details}

For both SMR prediction models $G$ and $Q$, we employ an EfficientNet-B4 pre-trained on ImageNet as $E(\cdot)$ to encode the original and compressed image into representations with $d=1792$. The input and output dimensions of FC layers in $R(\cdot)$ are $(1792\times2, 4096), (4096, 4096), \text{and } (4096, 1)$, respectively, with a ReLU module placed between neighboring FC layers. The input image resolutions are aligned with machines for different vision tasks, which are $224\times224$ for \textit{cls} and $512\times512$ for \textit{det}. We train the SMR models using the Adam optimizer with a learning rate of $10^{-4}$. To avoid any potential effects on an image's SMR, no data augmentation is used.

For \textit{det}, we find that some machines cannot have any detection on several original images after being filtered by $\mathcal{F}_{T_\text{conf}}$. Therefore, for subsequent experiments, we only use images with more than $20\%$ of machines having at least one detection after being filtered with $T_\text{conf}=0.3$. In our SMR dataset, $99.96\%$ of images meet this criterion. Notably, even if the requirement of proportion of machines increases to $90\%$, there are still $98.55\%$ of images satisfying it.

\subsection{Evaluation protocols} \label{subsec: evaluation methods and metrics}

The SMR prediction errors are straight given by the L1 loss during testing, \textit{i.e.} $|\Delta\text{SMR}|$. Additionally, to assess the potential for coding performance improvement using our SMR models, we consider real-world applications where the goal is to compress images and videos such that they achieve SMRs higher than a customizable threshold $T_{\text{SMR}}$. The intuitive method of attaining this target is to select a constant QP based on a known SMR distribution. However, this approach does not yield optimal results for every image, as many can tolerate higher QPs without violating the SMR target. Therefore, we use predicted SMRs to realize perceptual coding for general machines and improve the performance in the following steps:

\begin{enumerate}[leftmargin=14pt, noitemsep, topsep=0pt, label={\arabic*.}]
  \item We select several SMR thresholds to form a set $\mathbb{T}_\text{SMR}$ that covers a wide range of bit-rates and reachable SMR values. 
  \item For each threshold $T_{\text{SMR}}$ in $\mathbb{T}_\text{SMR}$, referring to a known SMR distribution $\mathcal{D}_\text{SMR}=\{\overline{\text{SMR}_{q_1}}, \overline{\text{SMR}_{q_2}}, \ldots, \overline{\text{SMR}_{q_n}}\}$, we search the QP $q_b$ satisfying $\overline{\text{SMR}_{q_b}} \geq T_{\text{SMR}}$ as the baseline. If no QP is exactly matched, we select the one with the minimum $|\overline{\text{SMR}_{q_b}}-T_{\text{SMR}}|$.
  \item We predict SMRs of images compressed with $\text{QP}=q_{b}, q_{b+1}, q_{b+2}, \ldots, q_n$ using the proposed SMR models. The predicted SMRs are denoted as $\text{SMR}_{\text{pred}}(\cdot)$.
  \item We search the predicted SMRs reversely to find the first $\text{SMR}_{\text{pred}}(I_{q_{b+k}}) \geq T_{\text{SMR}}$. The corresponding $q_{b+k}$ is considered the optimal QP for compressing $I_0$.
\end{enumerate}

Then, the average bit-rates in bits per pixel (bpp) and mean \textbf{actual} SMRs under all SMR thresholds are recorded. After that, the coding performance improvement is evaluated by Bjøntegaard Delta rate (BD-rate) \cite{bjontegaard2001calculation}, which corresponds to the average bit-rate difference \textbf{in percent} for the same SMR, the lower, the better. In cases where the original image has lower SMRs than $T_{\text{SMR}}$ across all QPs in the search range, $q_{b}$ is used for calculating BD-rate gains brought by ground truth SMRs, \textit{i.e.} $q_{b+k}=q_{b}$. Similarly, if the predicted SMRs are all lower than $T_{\text{SMR}}$ in the QP search range, $q_{b}$ is also applied for calculating BD-rate gains brought by predicted SMRs. Moreover, we compute the average MAE of QP $q_{b+k}$ obtained from using the ground truth and predicted SMR for all $T_{\text{SMR}}$, which is denoted as $|\Delta\text{QP}|$. The corresponding average MAE of PSNR is also recorded as $|\Delta\text{PSNR}|$. These two metrics are used to further assess the SMR prediction performance. Note that we exclude the case when $q_{b+k}$ is not found using the ground truth SMR during their calculations.

It is worth mentioning that SMR prediction is like a fine-grained image quality assessment task \cite{zhang2021fine} for machines, especially when evaluating performance within a broad range of adjacent QPs and closely matched bit-rates. This is far more challenging than a coarse-grained task that only considers a few widely spaced QPs or distant bit-rates \cite{zhang2021fine}, where quality differences are more pronounced and easier to distinguish.

\subsection{Basic results}

\begin{figure}[htbp]
  \centering
  \subfloat[Image classification]{
      \includegraphics[width=0.48\textwidth]{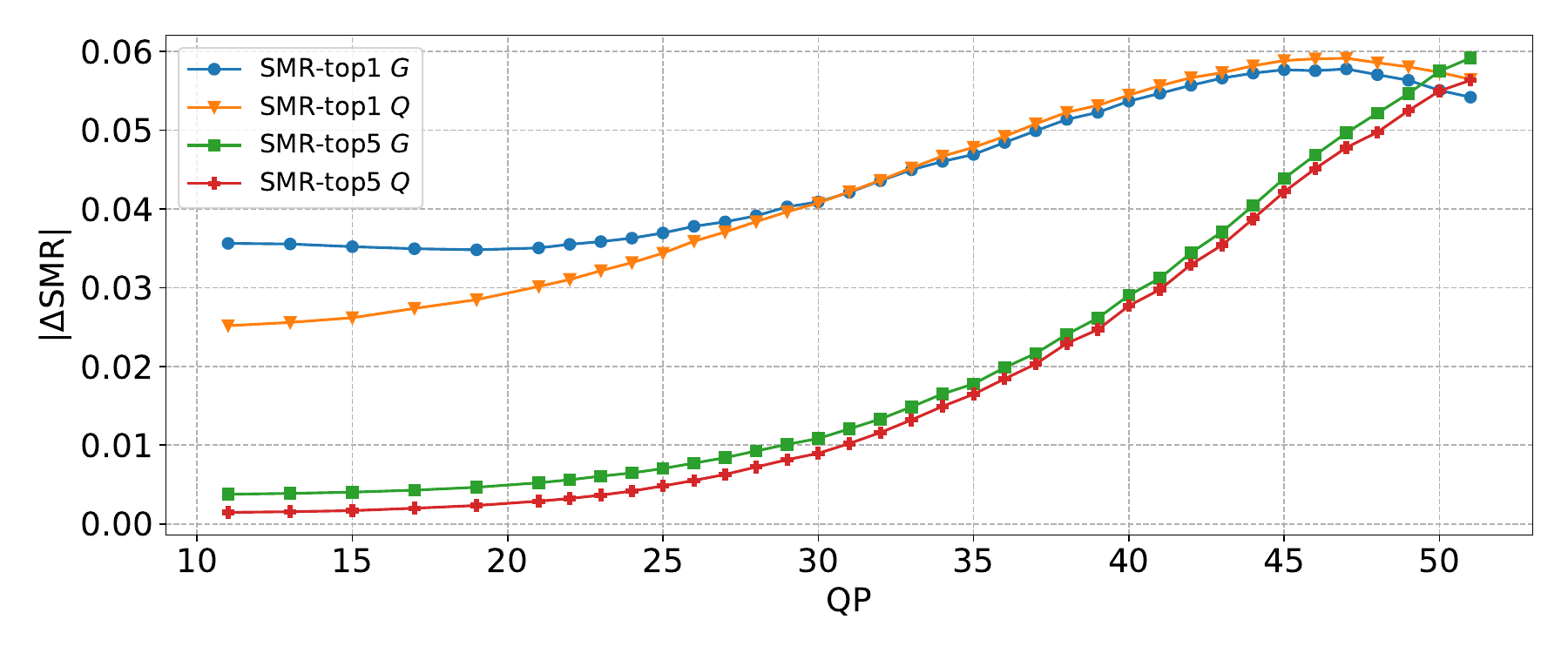}
  }
  \quad
  \subfloat[Object detection]{
      \includegraphics[width=0.48\textwidth]{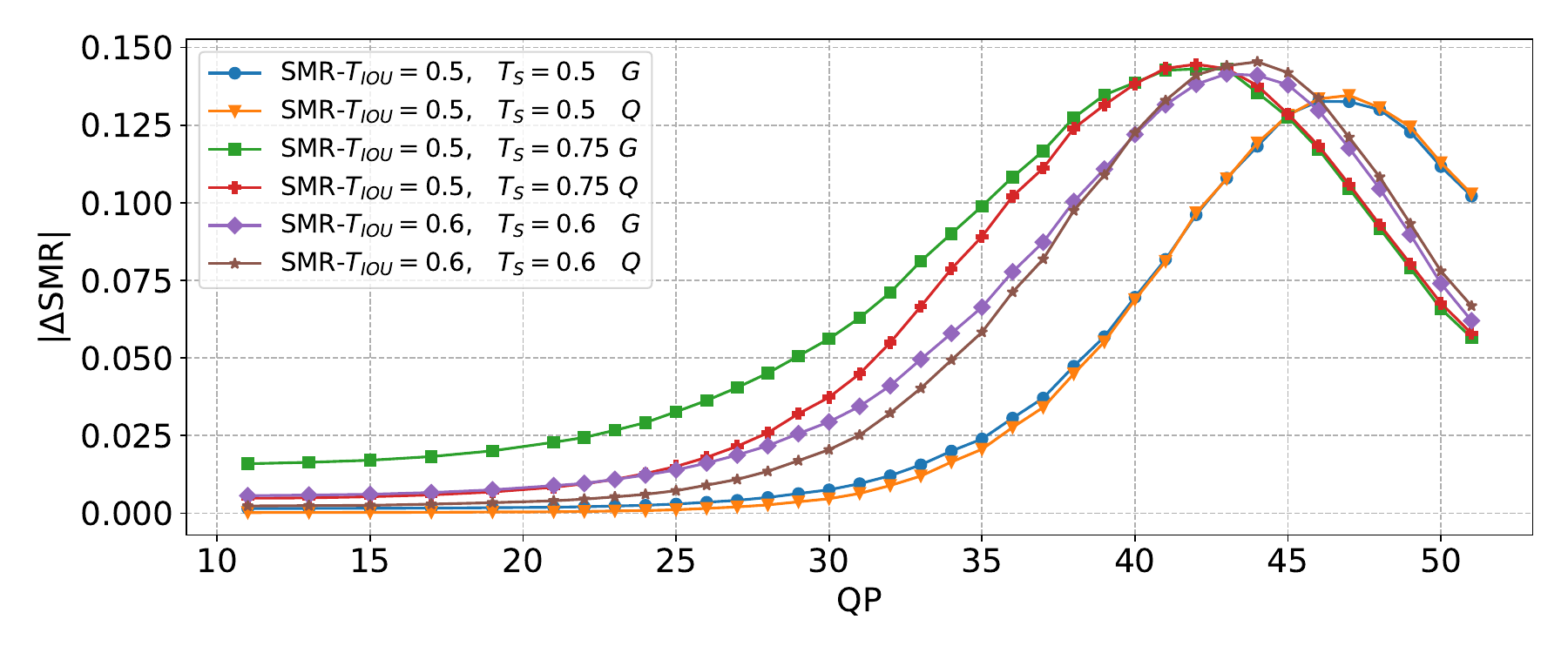}
  }
  \caption{The distribution of SMR prediction errors under different QPs.}
  \label{fig:qp-smr prediction distribution}
\end{figure}

\begin{figure*}[htbp]
  \subfloat{
      \includegraphics[width=0.19\textwidth]{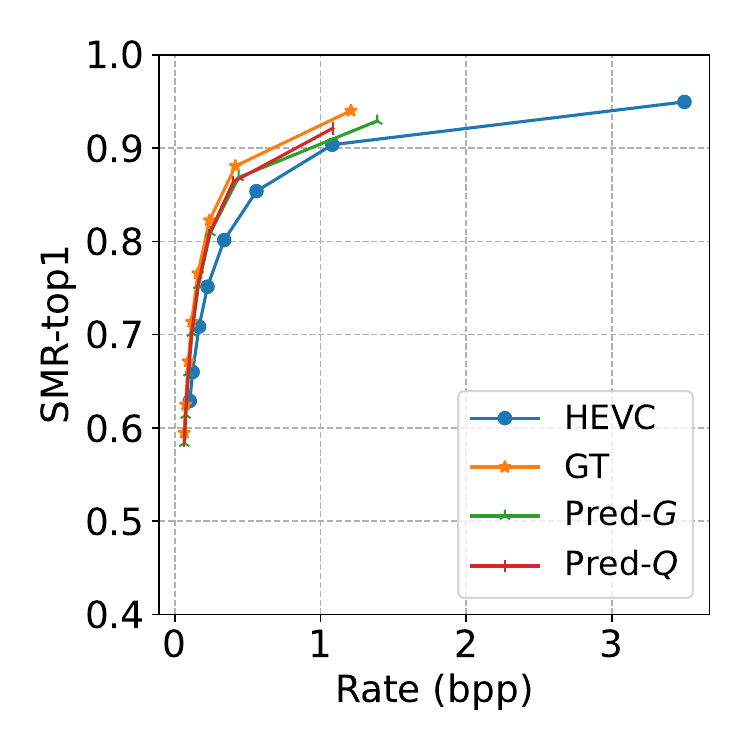}
  }
  \subfloat{
      \includegraphics[width=0.19\textwidth]{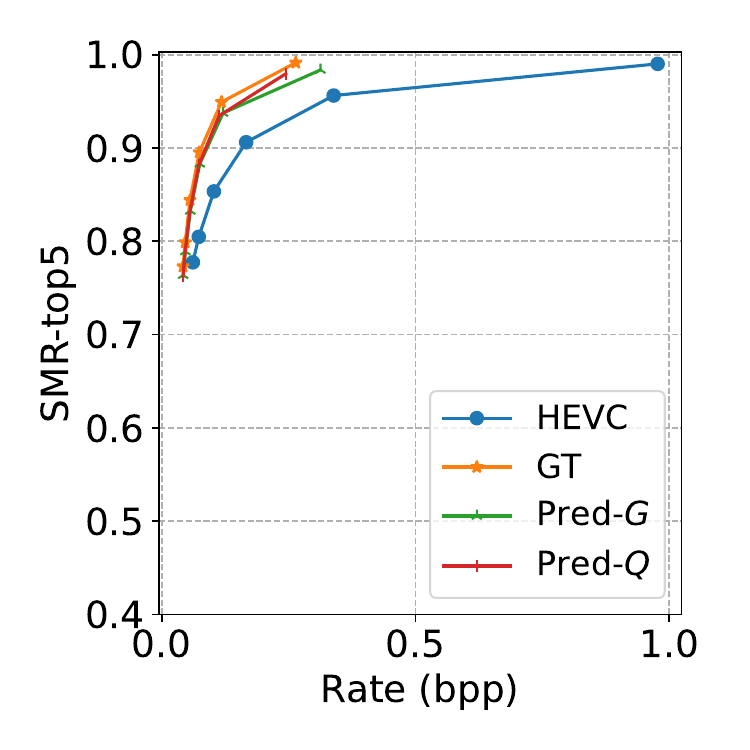}
  }
  \subfloat{
      \includegraphics[width=0.19\textwidth]{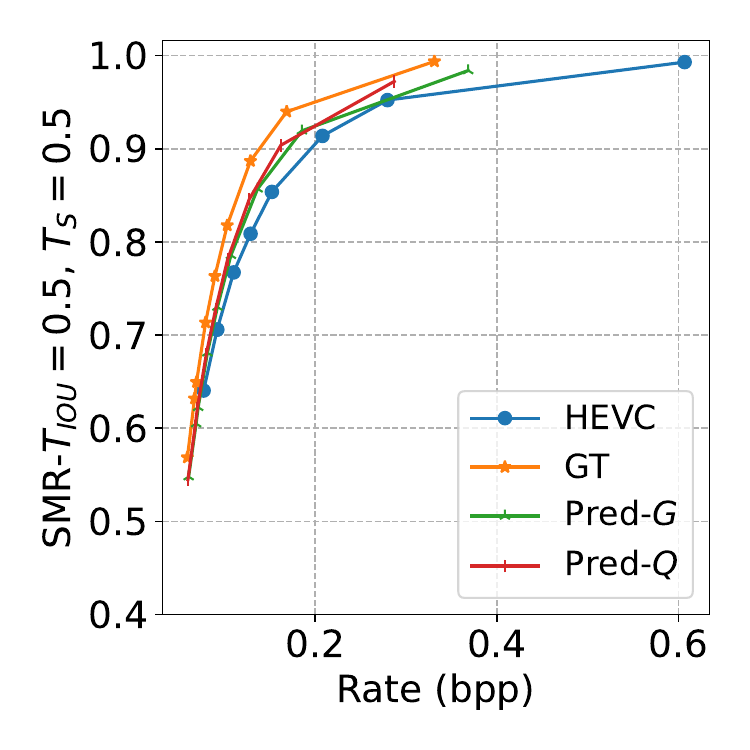}
  }
  \subfloat{
      \includegraphics[width=0.19\textwidth]{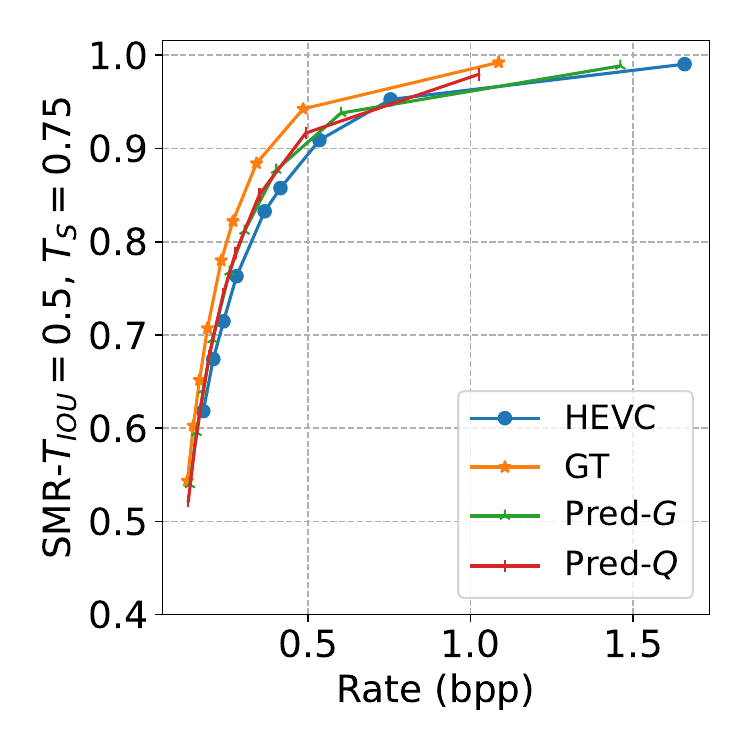}
  }
  \subfloat{
      \includegraphics[width=0.19\textwidth]{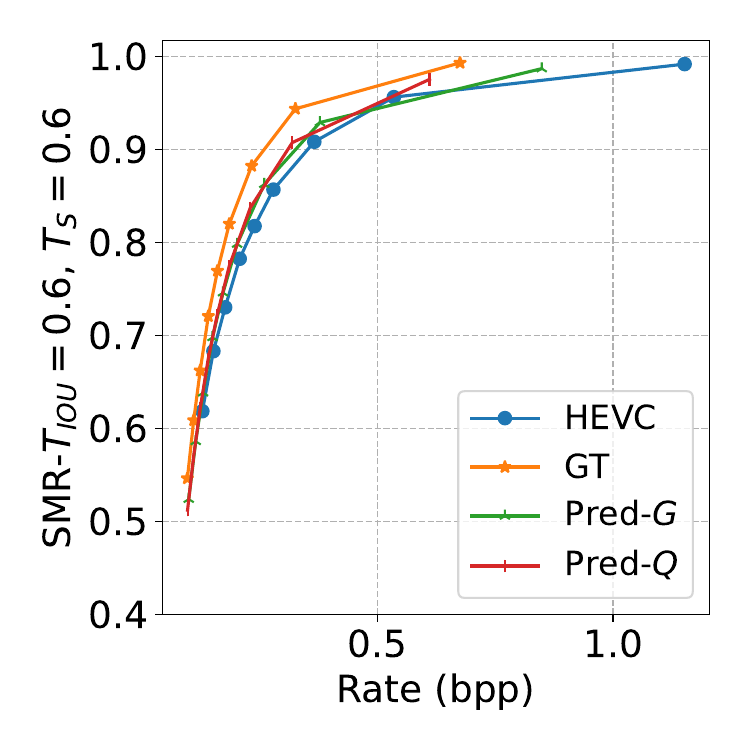}
  }
  \caption{Rate-SMR curves for all types of SMR of the two targeted tasks using ground truth SMR, predicted SMR from model $G$, and predicted SMR from model $Q$ to optimize QP selection, respectively.}
  \label{fig:rate-smr curves}
\end{figure*}

\bgroup
\def\arraystretch{1.2}
\begin{table*}[htbp]
  \centering
  \caption{Performances of SMR models for image classification.}
  \label{table:smr prediction results cls}
  \resizebox{\textwidth}{!}{
  \begin{tabular}{c|l|c|c|c|c|c|c|c|c|c|c|c|c|c|c}
  \Xhline{4\arrayrulewidth}
   & Dataset-Codec & \multicolumn{2}{c|}{COCO-HEVC} & \multicolumn{2}{c|}{COCO-VVC} & \multicolumn{2}{c|}{COCO-AVS3} & \multicolumn{2}{c|}{COCO-\cite{cheng2020learned}} & \multicolumn{2}{c|}{COCO-\cite{liu2023learned}} & \multicolumn{2}{c|}{VOC-HEVC} & \multicolumn{2}{c}{TVD-VVC} \\ \hline

   & Library & v1 & v2 & v1 & v2 & v1 & v2 & v1 & v2 & v1 & v2 & v1 & v2 & v1 & v2 \\
  \Xhline{3\arrayrulewidth}

  \multirow{9}{*}{SMR-top1} & $|\Delta\text{SMR}|$-$G$ & 0.0463 & 0.0454 & 0.0517 & 0.0505 & 0.0494 & 0.0549 & 0.0480 & 0.0466 & 0.0474 & 0.0461 & 0.0370 & 0.0468 & 0.0563 & 0.0549 \\

   & $|\Delta\text{SMR}|$-$Q$ & 0.0434 & 0.0417 & 0.0495 & 0.0474 & 0.0481 & 0.0569 & 0.0469 & 0.0443 & 0.0475 & 0.0449 & 0.0450 & 0.0593 & 0.0698 & 0.0685 \\ \cline{2-16}
   
   & $|\Delta\text{QP}|$-$G$ & 1.40 & 1.42 & 2.01 & 1.99 & 1.83 & 1.89 & 0.19 & 0.18 & 0.20 & 0.19 & 0.90 & 1.06 & 2.49 & 2.44 \\ 
   
   & $|\Delta\text{QP}|$-$Q$ & 1.22 & 1.20 & 1.85 & 1.79 & 1.60 & 1.78 & 0.16 & 0.16 & 0.17 & 0.16 & 0.82 & 1.03 & 2.27 & 2.19 \\ \cline{2-16}
   
   & $|\Delta\text{PSNR}|$-$G$ & 0.82 & 0.84 & 1.09 & 1.08 & 0.83 & 0.86 & 0.25 & 0.25 & 0.30 & 0.30 & 0.50 & 0.61 & 1.27 & 1.26 \\
   
   & $|\Delta\text{PSNR}|$-$Q$ & 0.69 & 0.67 & 0.98 & 0.93 & 0.70 & 0.79 & 0.22 & 0.21 & 0.27 & 0.26 & 0.44 & 0.58 & 1.16 & 1.13 \\ \cline{2-16}

   & \textbf{BD-rate GT} & \textbf{-39.2} & \textbf{-37.8} & \textbf{-34.0} & \textbf{-33.6} & \textbf{-41.2} & \textbf{-41.5} & \textbf{-15.5} & \textbf{-14.8} & \textbf{-15.4} & \textbf{-14.8} & \textbf{-38.4} & \textbf{-39.7} & \textbf{-32.5} & \textbf{-30.9} \\

   & BD-rate Pred-$G$ & -29.2 & -30.2 & -19.9 & -22.6 & -31.2 & -31.5 & -9.4 & -9.3 & -9.3 & -9.1 & -30.3 & -31.3 & -19.9 & -22.4 \\

   & BD-rate Pred-$Q$ & -30.4 & -31.4 & -21.3 & -24.1 & -32.4 & -32.9 & -13.4 & -13.4 & -13.2 & -13.3 & -32.7 & -34.6 & -21.6 & -24.4 \\

  \Xhline{3\arrayrulewidth}
  
  \multirow{9}{*}{SMR-top5} & $|\Delta\text{SMR}|$-$G$ & 0.0217 & 0.0211 & 0.0401 & 0.0390 & 0.0251 & 0.0404 & 0.0135 & 0.0130 & 0.0131 & 0.0126 & 0.0154 & 0.0355 & 0.0401 & 0.0388 \\

   & $|\Delta\text{SMR}|$-$Q$ & 0.0197 & 0.0189 & 0.0392 & 0.0381 & 0.0236 & 0.0421 & 0.0135 & 0.0129 & 0.0129 & 0.0123 & 0.0150 & 0.0367 & 0.0415 & 0.0411 \\ \cline{2-16} 
   
   & $|\Delta\text{QP}|$-$G$ & 1.03 & 0.99 & 1.97 & 1.92 & 1.26 & 0.77 & 0.13 & 0.12 & 0.13 & 0.13 & 0.48 & 0.91 & 2.40 & 2.30 \\ 
   
   & $|\Delta\text{QP}|$-$Q$ & 0.86 & 0.83 & 1.74 & 1.69 & 1.03 & 0.73 & 0.10 & 0.10 & 0.10 & 0.10 & 0.45 & 0.92 & 2.19 & 2.12 \\ \cline{2-16}
   
   & $|\Delta\text{PSNR}|$-$G$ & 0.53 & 0.51 & 0.91 & 0.89 & 0.50 & 0.28 & 0.17 & 0.17 & 0.20 & 0.20 & 0.23 & 0.48 & 1.19 & 1.14 \\
   
   & $|\Delta\text{PSNR}|$-$Q$ & 0.43 & 0.42 & 0.79 & 0.77 & 0.40 & 0.27 & 0.14 & 0.14 & 0.16 & 0.16 & 0.21 & 0.48 & 1.10 & 1.06 \\ \cline{2-16}

   & \textbf{BD-rate GT} & \textbf{-50.6} & \textbf{-48.5} & \textbf{-54.4} & \textbf{-54.2} & \textbf{-50.8} & \textbf{-49.5} & \textbf{-18.0} & \textbf{-15.8} & \textbf{-19.5} & \textbf{-17.3} & \textbf{-40.4} & \textbf{-48.2} & \textbf{-56.8} & \textbf{-56.0} \\

   & BD-rate Pred-$G$ & -43.4 & -42.0 & -44.2 & -45.3 & -43.1 & -45.0 & -9.0 & -8.9 & -9.2 & -9.1 & -35.4 & -39.9 & -44.7 & -46.0 \\

   & BD-rate Pred-$Q$ & -44.6 & -43.3 & -45.5 & -46.7 & -44.3 & -46.4 & -14.1 & -14.0 & -15.5 & -15.4 & -35.7 & -40.1 & -46.2 & -47.5 \\

  \Xhline{4\arrayrulewidth}
  \end{tabular}
  } 
\end{table*}
\egroup

As a demonstration, we select several types of SMR to train the proposed SMR models. Specifically, for \textit{cls}, we train two pairs of models $G$ and $Q$ (4 models in total) to predict SMR-top1 and SMR-top5 using the SMR labels annotated by the v1 machine library. For \textit{det}, we train three pairs of models $G$ and $Q$ to predict SMR with $(T_\text{IOU}, T_S)$ set to $(0.5, 0.5)$, $(0.5, 0.75)$, and $(0.6, 0.6)$ (6 models in total). Predicting for other SMR types should be similar. $|\Delta\text{SMR}|$ values of the two targeted tasks can be found in TABLE~\ref{table:smr prediction results cls} (3rd column) and \ref{table:smr prediction results det} (4th column), respectively, and the distributions of $|\Delta\text{SMR}|$ under different QPs are presented in Fig.~\ref{fig:qp-smr prediction distribution}. For \textit{cls}, an overall positive correlation exists between QP and $|\Delta\text{SMR}|$, suggesting higher prediction errors at higher QPs. For \textit{det}, such correlation becomes more non-monotonic, with the lowest prediction performance observed in the QP range of $[42, 47]$. The reason is that SMRs for \textit{det} are consistently low at high QPs, making prediction easier when SMR models detect significant differences between extracted representations from original and compressed images. However, when QPs are moderately high but not extreme, the feature differences are more ambiguous, leading to less accurate SMR predictions.

Then, we apply the methodology described in Section \ref{subsec: evaluation methods and metrics} to evaluate the coding performance optimized with SMR and calculate $|\Delta\text{QP}|$ and $|\Delta\text{PSNR}|$. For SMR-top1, $\mathbb{T}_\text{SMR}=[0.6:0.05:0.95]$, which means values range from 0.6 to 0.95 with a step size of 0.05. For SMR-top5, $\mathbb{T}_\text{SMR}=[0.75:0.05:0.95]$ plus $0.99$. For all types of SMR for \textit{det}, $\mathbb{T}_\text{SMR}=[0.6:0.05:0.95]$ plus $0.99$. The resulting rate-SMR curves are shown in Fig.~\ref{fig:rate-smr curves}, where ``HEVC'' is the non-optimized results by constant QPs, ``GT'' is the optimized results by applying ground truth SMRs, ``Pred $G$'' is by applying predicted SMRs from baseline models, and ``Pred $Q$'' is by applying predicted SMRs from SMR difference-based models. More precise results are presented in TABLE~\ref{table:smr prediction results cls} (3rd column) and \ref{table:smr prediction results det} (4th column).

\bgroup
\def\arraystretch{1.2}
\begin{table}[htbp]
  \centering
  \caption{Performances of SMR models for object detection.}
  \label{table:smr prediction results det}
  \resizebox{0.48\textwidth}{!}{
  \begin{tabular}{c|c|l|c|c|c|c|c}
  \Xhline{4\arrayrulewidth}
  $T_\text{IOU}$ & $T_S$ & Codec & HEVC & VVC & AVS3 & \cite{cheng2020learned} & \cite{liu2023learned} \\ 
  \Xhline{3\arrayrulewidth}
  \multirow{9}{*}{0.5} & \multirow{9}{*}{0.5} & $|\Delta\text{SMR}|$-$G$ & 0.0442 & 0.0672 & 0.0566 & 0.0159 & 0.0149 \\ 
   &  & $|\Delta\text{SMR}|$-$Q$ & 0.0429 & 0.0731 & 0.0583 & 0.0174 & 0.0155 \\ \cline{3-8} 
   &  & $|\Delta\text{QP}|$-$G$ & 1.68 & 2.25 & 2.23 & 0.21 & 0.19 \\ 
   &  & $|\Delta\text{QP}|$-$Q$ & 1.61 & 2.29 & 2.17 & 0.21 & 0.18 \\ \cline{3-8} 
   &  & $|\Delta\text{PSNR}|$-$G$ & 0.83 & 1.08 & 0.83 & 0.29 & 0.30 \\ 
   &  & $|\Delta\text{PSNR}|$-$Q$ & 0.79 & 1.11 & 0.80 & 0.29 & 0.28 \\ \cline{3-8} 
   &  & \textbf{BD-rate GT} & \textbf{-23.8} & \textbf{-25.0} & \textbf{-25.9} & \textbf{-21.9} & \textbf{-21.6} \\ 
   &  & BD-rate Pred-$G$ & -10.3 & -4.8 & -10.1 & -3.2 & -3.0 \\ 
   &  & BD-rate Pred-$Q$ & -12.3 & -4.9 & -11.4 & -3.3 & -2.9 \\
   
  \Xhline{3\arrayrulewidth}

  \multirow{9}{*}{0.5} & \multirow{9}{*}{0.75} & $|\Delta\text{SMR}|$-$G$ & 0.0730 & 0.0723 & 0.0872 & 0.0629 & 0.0586 \\ \cline{3-8} 
   &  & $|\Delta\text{SMR}|$-$Q$ & 0.0644 & 0.0804 & 0.0901 & 0.0775 & 0.0683 \\ \cline{3-8} 
   &  & $|\Delta\text{QP}|$-$G$ & 2.93 & 3.07 & 4.56 & 0.19 & 0.18 \\ 
   &  & $|\Delta\text{QP}|$-$Q$ & 2.45 & 2.95 & 3.61 & 0.17 & 0.15 \\ \cline{3-8} 
   &  & $|\Delta\text{PSNR}|$-$G$ & 1.72 & 1.76 & 2.19 & 0.26 & 0.28 \\ 
   &  & $|\Delta\text{PSNR}|$-$Q$ & 1.39 & 1.64 & 1.60 & 0.24 & 0.24 \\ \cline{3-8} 
   &  & \textbf{BD-rate GT} & \textbf{-23.2} & \textbf{-22.3} & \textbf{-25.8} & \textbf{-9.7 }& \textbf{-10.9} \\ 
   &  & BD-rate Pred-$G$ & -8.7 & -6.6 & -7.5 & -4.1 & -2.7 \\ 
   &  & BD-rate Pred-$Q$ & -10.3 & -4.7 & -7.9 & -1.6 & -2.1 \\

  \Xhline{3\arrayrulewidth}

  \multirow{9}{*}{0.6} & \multirow{9}{*}{0.6} & $|\Delta\text{SMR}|$-$G$ & 0.0599 & 0.0690 & 0.0779 & 0.0402 & 0.0374 \\ \cline{3-8} 
   &  & $|\Delta\text{SMR}|$-$Q$ & 0.0569 & 0.0804 & 0.0824 & 0.0473 & 0.0417 \\ \cline{3-8} 
   &  & $|\Delta\text{QP}|$-$G$ & 2.30 & 2.57 & 3.24 & 0.19 & 0.18 \\ 
   &  & $|\Delta\text{QP}|$-$Q$ & 2.06 & 2.65 & 3.01 & 0.16 & 0.14 \\ \cline{3-8} 
   &  & $|\Delta\text{PSNR}|$-$G$ & 1.27 & 1.39 & 1.36 & 0.25 & 0.29 \\ 
   &  & $|\Delta\text{PSNR}|$-$Q$ & 1.11 & 1.41 & 1.23 & 0.23 & 0.23 \\ \cline{3-8} 
   &  & \textbf{BD-rate GT} & \textbf{-23.2} & \textbf{-22.7} & \textbf{-25.5} & \textbf{-10.7} & \textbf{-12.4} \\ 
   &  & BD-rate Pred-$G$ & -8.4 & -3.2 & -6.6 & -3.9 & -2.6 \\
   &  & BD-rate Pred-$Q$ & -10.0 & -0.2 & -6.6 & -3.0 & -3.5 \\
   
  \Xhline{4\arrayrulewidth}
  \end{tabular}
  } 
\end{table}
\egroup

Several conclusions can be drawn from these results: (1) Using ground truth SMRs to determine appropriate QPs significantly improves coding performance for machines. For \textit{cls}, the BD-rate savings are $39.2\%$ and $50.6\%$ for SMR-top1 and SMR-top5, respectively. For \textit{det}, the BD-rate savings are $23.8\%$, $23.2\%$, and $23.2\%$ for three $(T_\text{IOU}, T_S)$ pairs, respectively. As a reference, the state-of-the-art (SOTA) coding standard VVC exceeds the last-generation standard HEVC by around $30\%$ \cite{zhao2020comparative}. (2) Using predicted SMRs also increases coding efficiency considerably, though not achieving the best performance due to prediction errors, demonstrating the effectiveness of our SMR models. Particularly for \textit{cls}, BD-rate savings are approximately $30\%$ for SMR-top1 and even more impressive at $43\%$ for SMR-top5. These gains are cross-generational, underscoring the potential of our perceptual coding approach for machines. (3) The accuracy in predicting adequate QP using SMR models is generally satisfactory as $|\Delta\text{QP}|$ and $|\Delta\text{PSNR}|$ are low. However, the performance varies for different tasks and SMR types. Simpler task or less strict judgment of machine perception consistency leads to higher prediction accuracy. For instance, \textit{cls}$>$\textit{det}, and SMR-top5$>$SMR-top1. (4) Models $Q$ consistently outperform $G$ by achieving lower prediction error and higher coding gains, showing its enhanced capability in SMR prediction. (5) The optimal/practical coding gains achieved with ground truth/predicted SMR vary by task. Greater gains are observed in \textit{cls} than \textit{det} as more images can preserve a high SMR for \textit{cls} after heavy compression. This allows for selecting higher QPs more frequently during optimization, resulting in more bit savings while still satisfying enough machines. (6) Coding gains differ among SMR types due to different SMR prediction performances and distribution characteristics. For \textit{cls}, more coding gains can be achieved for SMR-top5 than SMR-top1 because of similar reasons in (3)/(5), coupled with a lower $|\Delta\text{SMR}|$. For \textit{det}, the same trend is observed.

\subsection{Generalize on unseen machines} \label{subsec: generalize to more machines}

We design SMR models to learn representative features that implicitly capture the perceptual characteristics of general machines. They have the potential to generalize on more unseen machines. To verify such generalizability, we keep our SMR models trained on the v1 machine library and test them on the v2 library, where the additional 14 machines are unknown to them. Experimental results are presented in TABLE~\ref{table:smr prediction results cls} (4th column), which indicate comparable SMR prediction errors and coding gains on the v2 library. Importantly, these results are also consistent in the following experiments, which will be described in later sections. Because the differences between two machine libraries in both ground truth (as shown in Section \ref{subsec:dataset study}) and predicted SMRs are slight, the proposed SMR models can adapt to many real-world applications where diverse machines are used, even unseen ones.


\subsection{Generalize on unseen codecs} \label{subsec: generalize to unseen codecs}

An ideal SMR model should generalize well on different codecs used in various applications. To assess this aspect, we evaluate the proposed SMR models on a range of unseen codecs, including traditional, neural, and MVS-oriented codecs. For traditional codecs, we select VVC and AVS3 as they are two SOTAs. Their reference softwares, VTM-20.0 \cite{vtm} and HPM-15.1 \cite{hpm}, are used for the compression. It is worth mentioning that VTM-20.0 is used as the inner codec in the MPEG VCM standard \cite{vcm-ctc,vcm-rs}. To produce even more differences, we increase the internal bit depth to 10 for VVC (as a comparison, both HEVC and AVS3 are using the 8-bit compression configuration). Since the QP range is different between HEVC and these two codecs, we re-select QPs as $11, 16, 21, 24, 27, 30, 32, 34, 36, 37, \dots, 63$ to cover a large bit-rate range. For each codec, $q_b$ for the corresponding $T_\text{SMR}$ in $\mathbb{T}_\text{SMR}$ is re-determined. Furthermore, the evaluation also extends to the test dataset annotated by the v2 machine library for \textit{cls}, presenting an even greater challenge to the SMR prediction models as they are still trained on the HEVC-compressed v1 train dataset.

\begin{figure}[htbp]
  \centering
  \subfloat[Image classification]{
      \includegraphics[width=0.23\textwidth]{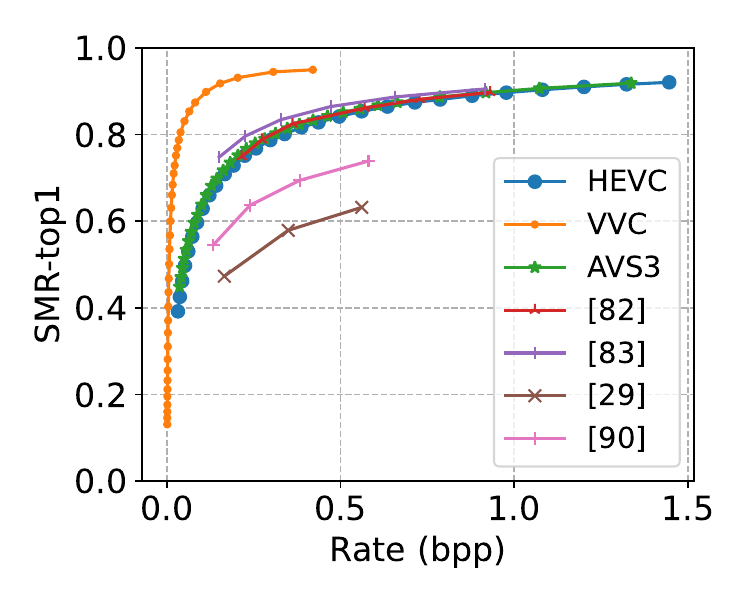}
  }
  \subfloat[Object detection]{
      \includegraphics[width=0.23\textwidth]{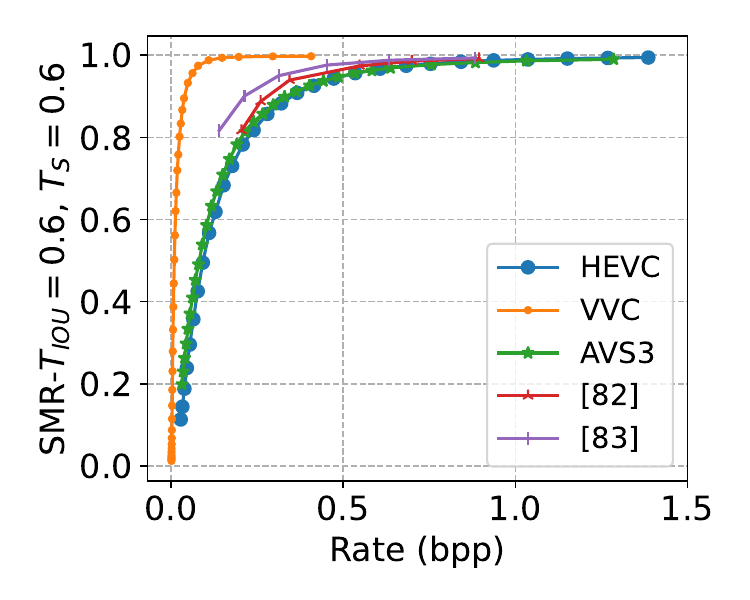}
  }
  \caption{Rate-SMR distributions of different codecs for different tasks. For HEVC and AVS3, the rate range can be extended to 3.0, but the SMR values after bpp=1.5 are so close that we cut the range to 1.5 for better visualization.}
  \label{fig:rate-smr different codecs}
\end{figure}

Results on VVC- and AVS3-compressed test datasets are detailed in TABLE~\ref{table:smr prediction results cls} (5-8th column) and \ref{table:smr prediction results det} (5-6th column). It can be observed that: (1) $|\Delta\text{SMR}|$, $|\Delta\text{QP}|$, and $|\Delta\text{PSNR}|$ all increase when applied to unseen codecs for both \textit{cls} and \textit{det}, while the errors remain within an acceptable range. (2) Models $Q$ cannot always outperform model $G$, particularly for \textit{det}. (3) Leveraging ground truth SMRs for optimal QP selection still achieves impressive coding gains for machines like HEVC. (4) Using predicted SMRs also improves coding performance consistently across different tasks, codecs, and machine library versions, demonstrating the good generalizability of proposed SMR models. However, coding gains on unseen codecs are relatively modest compared to HEVC, particularly on VVC, due to its unique SMR distribution. As depicted in Fig.~\ref{fig:rate-smr different codecs}, VVC requires much lower bpp to obtain high SMR value, and similar bit-rates can correspond to vastly different SMR values. Hence, inaccurate SMR predictions on VVC result in more substantial penalties than other codecs when used to improve coding efficiency. Nevertheless, as two of the most advanced traditional codecs, VVC and AVS3 significantly outperform HEVC \cite{zhao2020comparative}, yet our SMR models can help them further save a considerable amount of bits for machines. Therefore, SMR can be a practical solution to optimize currently used codecs for general machines, mitigating the need for costly and risky codec replacements.

Given the growing interest in neural codecs, we select two representative methods \cite{cheng2020learned,liu2023learned} to further evaluate the generalizability of SMR models, where \cite{cheng2020learned} is a well-known baseline and \cite{liu2023learned} is the SOTA. Note that these neural codecs offer much fewer compression operation points than traditional codecs, specifically limited to six ``QPs'' each. Moreover, their SMR ranges are also different, as illustrated in Fig.~\ref{fig:rate-smr different codecs}. Hence, we adapt our $T_\text{SMR}$ selections to align with their SMR distributions. For SMR-top1, $\mathbb{T}_\text{SMR}=[0.78, 0.82, 0.85, 0.88]$. For SMR-top5, $\mathbb{T}_\text{SMR}=[0.94, 0.96, 0.98, 0.99]$. For \textit{det}, $\mathbb{T}_\text{SMR}=[0.96, 0.98, 0.99, 1.0]$ when $(T_\text{IOU}, T_S)=(0.5, 0.5)$, $\mathbb{T}_\text{SMR}=[0.85, 0.9, 0.95, 0.98]$ when $(T_\text{IOU}, T_S)=(0.5, 0.75)$, and $\mathbb{T}_\text{SMR}=[0.93, 0.96, 0.98, 0.99]$ when $(T_\text{IOU}, T_S)=(0.6, 0.6)$. 

TABLE~\ref{table:smr prediction results cls} (9-12th column) and \ref{table:smr prediction results det} (7-8th column) presents the prediction performances and coding gains of SMR models on neural codecs. The results show that SMR prediction performances are comparable to those on traditional codecs. However, the maximum achievable coding gains using ground truth SMRs are constrained because of the more uniform SMR distributions, reduced number of operation points, and narrower rate ranges of neural codecs. These factors also lead to diminished coding gains when applying predicted SMRs. Meanwhile, models $G$ and $Q$ exhibit varying degrees of effectiveness across different scenarios. However, they still have good generalizability on unseen neural codecs, which is especially commendable considering the marked differences in compression distortions between traditional and neural codecs.

\bgroup
\def\arraystretch{1.28}
\begin{table}[htbp]
  \centering
  \caption{Performances of SMR models on MVS-oriented codecs.}
  \label{table:smr performance coco vcm}
  \resizebox{0.48\textwidth}{!}{
  \begin{tabular}{l|c|c|c|c|c|c|c|c}
  \Xhline{4\arrayrulewidth}
  Codec & \multicolumn{4}{c|}{\cite{bai2022towards}} & \multicolumn{4}{c}{\cite{chen2023transtic}} \\ \hline

  SMR type & top1-v1 & top1-v2 & top5-v1 & top5-v2 & top1-v1 & top1-v2 & top5-v1 & top5-v2 \\
  \Xhline{3\arrayrulewidth}

  $|\Delta\text{SMR}|$-$G$ & 0.0633 & 0.0617 & 0.0387 & 0.0375 & 0.0654 & 0.0633 & 0.0338 & 0.0325 \\

  $|\Delta\text{SMR}|$-$Q$ & 0.0541 & 0.0521 & 0.0404 & 0.0388 & 0.1359 & 0.1320 & 0.0379 & 0.0364 \\ \cline{1-9}
   
  PLCC-$G$ & 0.9615 & 0.9630 & 0.9517 & 0.9535 & 0.9495 & 0.9514 & 0.9353 & 0.9380 \\ 
   
  PLCC-$Q$ & 0.9691 & 0.9704 & 0.9494 & 0.9516 & 0.8135 & 0.8166 & 0.9246 & 0.9275 \\
   
  PLCC-LPIPS & 0.5688 & 0.5646 & 0.4302 & 0.4257 & 0.5176 & 0.5141 & 0.4061 & 0.4013 \\

  PLCC-DISTS & 0.6360 & 0.6327 & 0.4899 & 0.4863 & 0.5823 & 0.5796 & 0.4527 & 0.4489 \\

  PLCC-TOPIQ & 0.5408 & 0.5357 & 0.3736 & 0.3688 & 0.4858 & 0.4814 & 0.3396 & 0.3351 \\ \cline{1-9}
   
  SROCC-$G$ & 0.9011 & 0.9010 & 0.9078 & 0.9104 & 0.8861 & 0.8860 & 0.8645 & 0.8674 \\ 
   
  SROCC-$Q$ & 0.9660 & 0.9668 & 0.9220 & 0.9241 & 0.8968 & 0.8977 & 0.8785 & 0.8803 \\
   
  SROCC-LPIPS & 0.6470 & 0.6490 & 0.5608 & 0.5636 & 0.5976 & 0.5996 & 0.5176 & 0.5198 \\

  SROCC-DISTS & 0.7340 & 0.7369 & 0.6599 & 0.6641 & 0.6758 & 0.6788 & 0.6007 & 0.6045 \\

  SROCC-TOPIQ & 0.5884 & 0.5905 & 0.4956 & 0.4986 & 0.5613 & 0.5635 & 0.4767 & 0.4794 \\ \cline{1-9}

  \Xhline{4\arrayrulewidth}
  \end{tabular}
  } 
\end{table}
\egroup

Finally, we evaluate our SMR models on two SOTA MVS-oriented codecs \cite{bai2022towards,chen2023transtic} for \textit{cls}. Since they only offer 3 and 4 QPs that precludes the possibility of SMR-guided coding optimization, we mainly report $|\Delta\text{SMR}|$ results. Additionally, we calculate the Pearson Linear Correlation Coefficient (PLCC) and Spearman Rank Order Correlation Coefficient (SROCC) to assess the correlation between predicted and ground truth SMRs. These results are then compared with three full-reference image quality assessment (IQA) metrics, which are LPIPS \cite{zhang2018unreasonable}, DISTS \cite{ding2020image}, and TOPIQ \cite{chen2023topiq}. We include these IQA metrics for comparison due to their development based on deep feature differences, akin to our SMR models, and the fact that SMR itself is quality-related. 

The SMR distributions of these MVS-oriented codecs are shown in Fig.~\ref{fig:rate-smr different codecs}. Importantly, their SMRs are much lower than other codecs, as they have been optimized just for a specific machine, demonstrating the importance of our work. Nevertheless, the proposed SMR models have commendable generalization capabilities, as evidenced in TABLE~\ref{table:smr performance coco vcm}. Although model $Q$ shows relatively weaker performance, particularly for predicting SMR-top1 on \cite{chen2023transtic}, model $G$ achieves robust results across all SMR types for both codecs. Furthermore, our SMR models significantly outperform all IQA models in correlation. 

Based on all previous results, the proposed SMR models have broad and practical generalizability across various unseen codecs, including traditional, neural, and both HVS- and MVS-oriented ones. This achievement is particularly notable as the models are trained exclusively on the HEVC-compressed dataset. We anticipate that re-training the models to adapt to different codecs could further improve their performance. Meanwhile, for any codec, the performances of SMR models are similar between v1 and v2 machine libraries, again demonstrating the conclusions in Section \ref{subsec: generalize to more machines}.

\bgroup
\def\arraystretch{1.2}
\begin{table*}[htbp]
  \centering
  \caption{Performances of SMR models on inter frames.}
  \label{table:smr performance sfu}
  \resizebox{\textwidth}{!}{
  \begin{tabular}{l|c|c|c|c|c|c|c|c|c|c|c|c|c|c}
  \Xhline{4\arrayrulewidth}
  Codec-Config & \multicolumn{7}{c|}{VVC-RA} & \multicolumn{7}{c}{VVC-LD} \\ \hline

  SMR type & top1-v1 & top1-v2 & top5-v1 & top5-v2 & (0.5, 0.5) & (0.5, 0.75) & (0.6, 0.6) & top1-v1 & top1-v2 & top5-v1 & top5-v2 & (0.5, 0.5) & (0.5, 0.75) & (0.6, 0.6) \\
  \Xhline{3\arrayrulewidth}

  $|\Delta\text{SMR}|$-$G$ & 0.0553 & 0.0537 & 0.0353 & 0.0341 & 0.0698 & 0.0775 & 0.0839 & 0.0559 & 0.0542 & 0.0412 & 0.0399 & 0.0678 & 0.0708 & 0.0780 \\

  $|\Delta\text{SMR}|$-$Q$ & 0.0542 & 0.0525 & 0.0338 & 0.0327 & 0.0885 & 0.1132 & 0.1072 & 0.0551 & 0.0534 & 0.0413 & 0.0403 & 0.0896 & 0.1022 & 0.1012 \\ \cline{1-15}
   
  PLCC-$G$ & 0.9681 & 0.9696 & 0.9605 & 0.9624 & 0.9538 & 0.9489 & 0.9380 & 0.9725 & 0.9738 & 0.9647 & 0.9667 & 0.9602 & 0.9538 & 0.9467 \\ 
   
  PLCC-$Q$ & 0.9694 & 0.9710 & 0.9639 & 0.9657 & 0.9466 & 0.9349 & 0.9351 & 0.9732 & 0.9747 & 0.9655 & 0.9674 & 0.9527 & 0.9423 & 0.9426 \\
   
  PLCC-LPIPS & 0.6464 & 0.6486 & 0.5811 & 0.5820 & 0.9098 & 0.8600 & 0.8854 & 0.6941 & 0.6959 & 0.6344 & 0.6363 & 0.9120 & 0.8507 & 0.8808 \\

  PLCC-DISTS & 0.6827 & 0.6859 & 0.5980 & 0.6004 & 0.8796 & 0.8669 & 0.8723 & 0.7219 & 0.7249 & 0.6450 & 0.6494 & 0.8737 & 0.8557 & 0.8635 \\

  PLCC-TOPIQ & 0.5343 & 0.5333 & 0.4343 & 0.4320 & 0.8339 & 0.9018 & 0.8935 & 0.5676 & 0.5666 & 0.4683 & 0.4668 & 0.8572 & 0.9115 & 0.9083 \\ \cline{1-15}
   
  SROCC-$G$ & 0.9590 & 0.9607 & 0.9184 & 0.9228 & 0.9529 & 0.9499 & 0.9367 & 0.9669 & 0.9683 & 0.9414 & 0.9450 & 0.9618 & 0.9504 & 0.9461 \\ 
   
  SROCC-$Q$ & 0.9644 & 0.9662 & 0.9275 & 0.9311 & 0.9473 & 0.9445 & 0.9460 & 0.9705 & 0.9720 & 0.9472 & 0.9503 & 0.9565 & 0.9411 & 0.9506 \\
   
  SROCC-LPIPS & 0.6459 & 0.6473 & 0.6743 & 0.6760 & 0.9347 & 0.9324 & 0.9364 & 0.6998 & 0.7014 & 0.7221 & 0.7243 & 0.9434 & 0.9317 & 0.9407 \\

  SROCC-DISTS & 0.6912 & 0.6927 & 0.7134 & 0.7162 & 0.9302 & 0.9261 & 0.9244 & 0.7411 & 0.7433 & 0.7572 & 0.7608 & 0.9317 & 0.9220 & 0.9253 \\

  SROCC-TOPIQ & 0.5925 & 0.5923 & 0.6274 & 0.6281 & 0.9138 & 0.9095 & 0.9223 & 0.6242 & 0.6241 & 0.6543 & 0.6552 & 0.9226 & 0.9119 & 0.9262 \\ \cline{1-15}

  \Xhline{4\arrayrulewidth}
  \end{tabular}
  } 
\end{table*}
\egroup

\subsection{Generalize on unseen datasets} \label{subsec: generalize to other datasets}

We evaluate the capability of SMR models on two other datasets, which are the test2007 dataset of PASCAL VOC \cite{everingham2010pascal} and the TVD dataset \cite{gao2022open}, comprising 13315 and 1098 object images, respectively. It is worth mentioning that (i) most object categories of VOC are also included in COCO except ``sofa'', thereby the semantics of images on these two datasets differ slightly, and (ii) TVD is a test dataset in CTC of MPEG VCM standard \cite{vcm-ctc}. To introduce more variables, we compress VOC with HEVC and TVD with VVC, and we also test the models on v1- and v2-annotated datasets separately. For VOC, we adjust the last $T_\text{SMR}$ from $0.99$ to $0.97$, in line with its SMR distribution. The SMR prediction performances and coding gains for \textit{cls} are presented in TABLE~\ref{table:smr prediction results cls} (last 4 columns). According to the results, again, SMR models bring significant coding gains, model $Q$ outperforms $G$, and the performances are close between the v1 and v2 machine library, all remaining consistent with previous results on the COCO dataset compressed with traditional codecs. Therefore, our SMR models show strong generalizability on unseen datasets.

\subsection{Generalize on inter frames}

Like intra coding, inter coding is an essential part of video compression that significantly influences final quality. To evaluate the generalizability of SMR models on inter frames, we align with the CTC of MPEG VCM standard \cite{vcm-ctc} to compress SFU-HW dataset \cite{choi2021dataset} using random-access (RA) and low-delay (LD) configurations of VVC. This dataset contains 1165 video frames in multiple resolutions and 13732 objects. The same 36 QPs of VVC in Section \ref{subsec: generalize to unseen codecs} are used for compression. Since the inter frames are not compressed independently, we cannot directly select the optimal QP for each frame according to their SMRs. Therefore, we adopt the same evaluation methods as the experiment on MVS-oriented codecs in Section \ref{subsec: generalize to unseen codecs}. The results are presented in TABLE~\ref{table:smr performance sfu}, where the first four columns for each coding configuration are the results for \textit{cls}, and the last three are for \textit{det} in different $(T_\text{IOU}, T_S)$ pairs. The prediction performances of SMR models are close to intra frames. The prediction accuracies are still similar between two machine libraries. Model $Q$ shows weaker generalizability, especially for \textit{det}, while model $G$ consistently performs well on all tasks and SMR types. Moreover, our SMR models exhibit high correlation scores and surpass all IQA models, indicating the strong correlation between predicted and ground truth SMRs. Therefore, we can conclude that the proposed SMR models generalize well on inter frames. Exploring how to use SMR models to optimize inter coding efficiency for machines is worth studying in future works.

\subsection{Compare with state-of-the-art perceptual VCM methods}

We compare our methods with the SOTA perceptual VCM methods \cite{zhang2021just,zhang2023learning}, \textit{i.e.} Just Recognizable Distortion (JRD) models. These models actually find the first JND for a machine as described in Section \ref{subsec:smr vs jnd}. In this experiment, we manually locate ground truth JRDs of all 58/98 machines for \textit{cls}/\textit{det} to compare with our SMR models. We adopt two strategies to locate JRD: JRD Front and JRD back. For JRD Front, we search for the minimum QP that leads to PC variation in ascending order and denote the previous QP as JRD. For JRD Back, we search for the maximum QP where the machine can still keep perception the same as on the original image in descending order and denote it as JRD. Then, we compress the SMR test dataset with these QPs and record the bit-rate and SMR. For SMR models, we set $\mathbb{T}_\text{SMR}=[0.3:0.05:0.95]$ and $[0.2:0.05:0.95]$ to cover comparable bit-rate ranges for \textit{cls} and \textit{det}, respectively. All QPs serve as the search range for both SMR and JRD, defaulting to QP 11 if no QP meets the search criteria. 

\begin{figure}[htbp]
  \centering
  \subfloat[Image classification]{
      \includegraphics[width=0.23\textwidth]{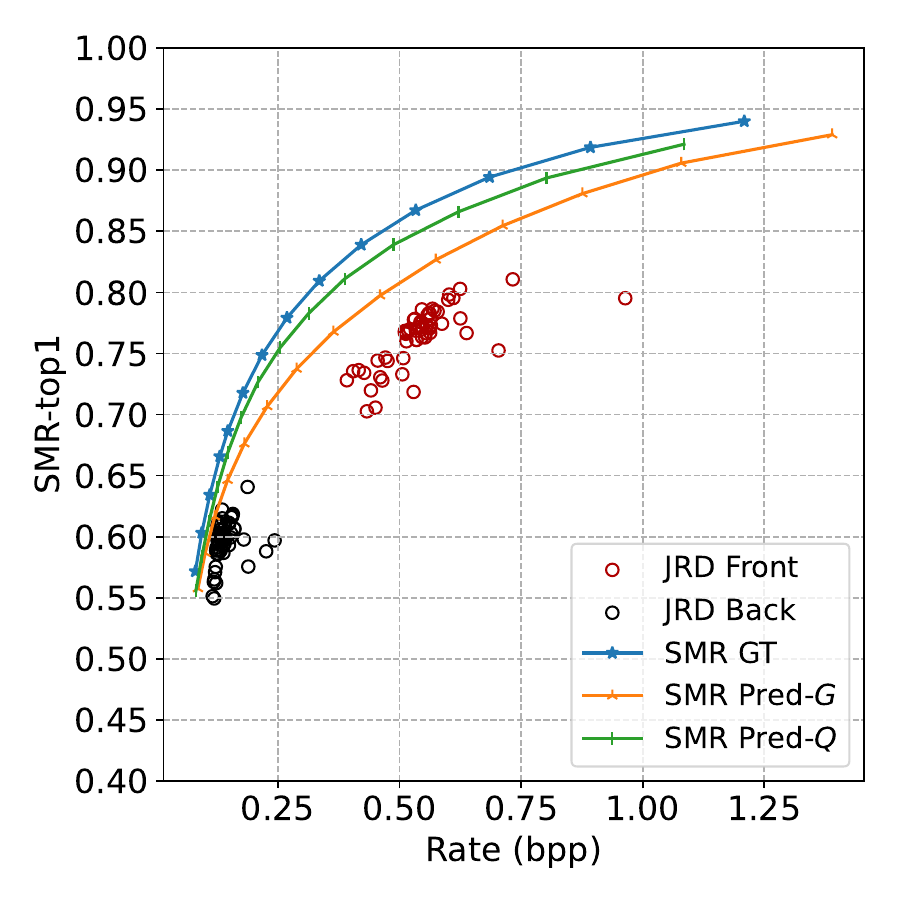}
  }
  \subfloat[Object detection]{
      \includegraphics[width=0.23\textwidth]{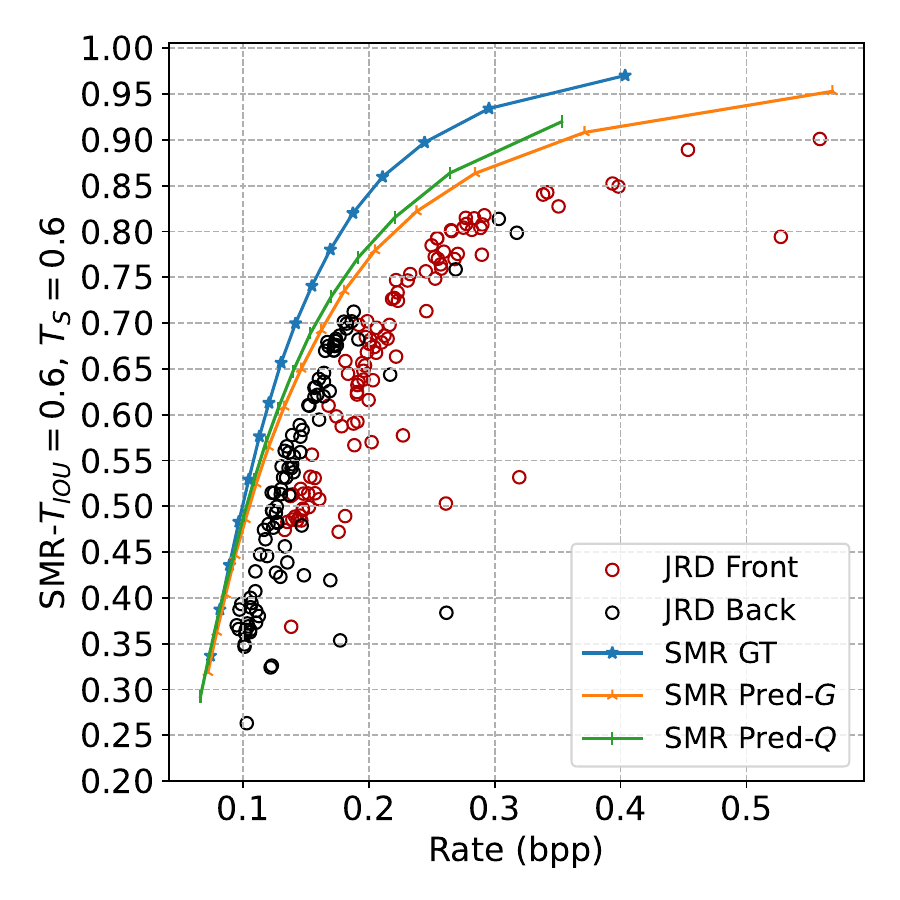}
  }
  \caption{Performance comparison between SMR and JRD.}
  \label{fig:smr vs jrd}
\end{figure}

The resulting rate-SMR curves are presented in Fig.~\ref{fig:smr vs jrd}. Notably, JRDs yield much lower SMR values than SMR models, as they optimize coding performance just for the corresponding single machine rather than general machines. Even we introduce JRD Front to force using lower QPs more frequently, which should lead to higher SMR in most cases, the rate-SMR performance is still inferior due to machine diversity. On the other hand, JRD Back can only achieve low SMR at low bit-rates, which is inapplicable in real-world scenarios. Therefore, SMR is preferred over JRD for perceptual VCM for diverse machines.

\section{Conclusion} \label{sec:conclusion}

This paper proposes a novel concept for VCM called Satisfied Machine Ratio (SMR). SMR statistically measures the quality of compressed images and videos for machines by modeling general machine perceptual characteristics. Targeting the image classification and object detection tasks, we build two machine libraries with up to 72 and 98 representative machines to study the general MVS behavior and create a large-scale SMR dataset with over 27 million images to facilitate SMR studies. Furthermore, we propose an SMR model based on the correlation between deep feature differences and SMR, which can predict the SMR of any compressed image. To leverage all labeled data and increase prediction accuracy, we propose another SMR model based on SMR differences between images in different compression quality levels. Extensive experiments prove the effectiveness of our proposed SMR models, revealing significant compression performance improvements for machines. Importantly, our SMR models have strong generalizability on unseen machines, codecs, datasets, and frame types. SMR enables perceptual coding for machines and propels VCM from specificity to generality. The idea and statistical methodology of considering general machines instead of a single specific machine can also benefit other related research areas.

\section*{Acknowledgments}

This work was supported in part by the National Natural Science Foundation of China under grant 62072008 and U20A20184, and the High Performance Computing Platform of Peking University, which are gratefully acknowledged.


\bibliographystyle{IEEEtran}
\bibliography{refs}

 


\begin{IEEEbiography}[{\includegraphics[width=1in,height=1.25in,clip,keepaspectratio]{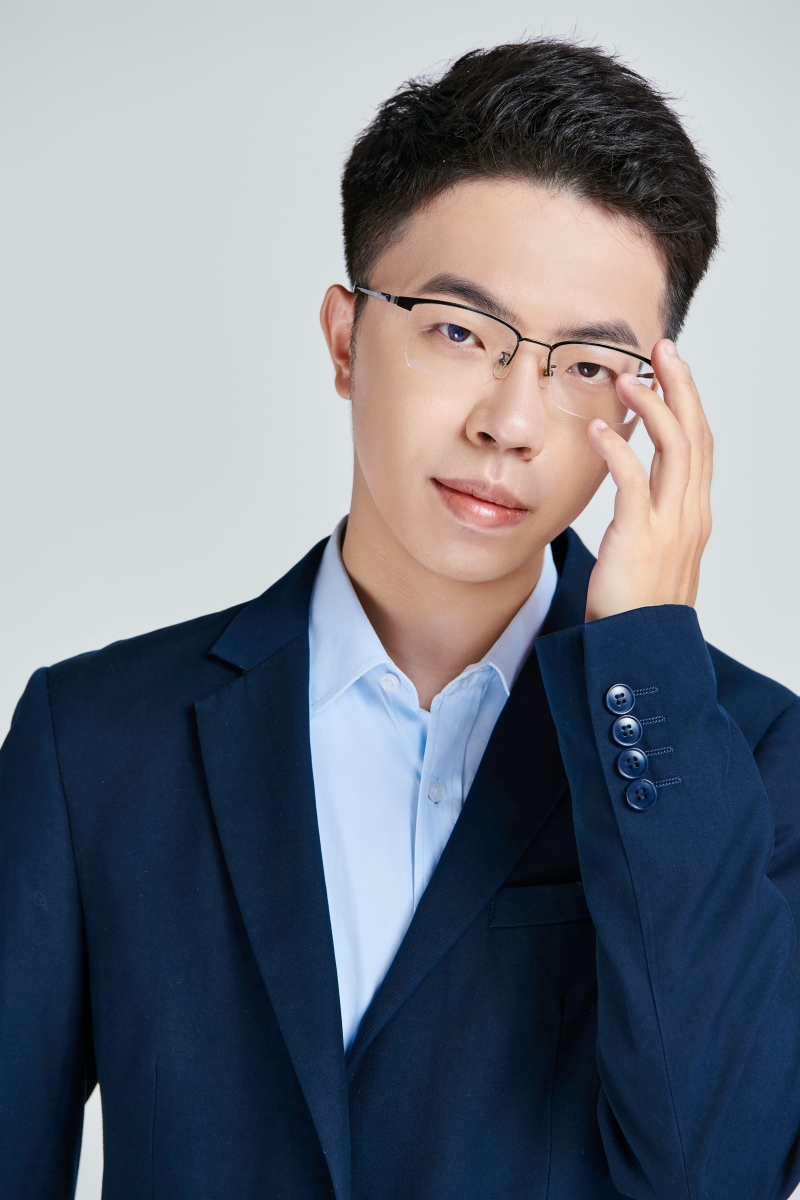}}]{Qi Zhang}
  (Student Member, IEEE) received the B.S. degree in software engineering from Hefei University of Technology, Hefei, China, in 2017, and the M.S. degree in computer application technology from Peking University, Beijing, China, in 2020. He is currently pursuing the Ph.D. degree with the Department of Computer Science, Peking University, Beijing. His research interests include video coding for both humans and machines, immersive video technology, and virtual reality.
\end{IEEEbiography}

\begin{IEEEbiography}[{\includegraphics[width=1in,height=1.25in,clip,keepaspectratio]{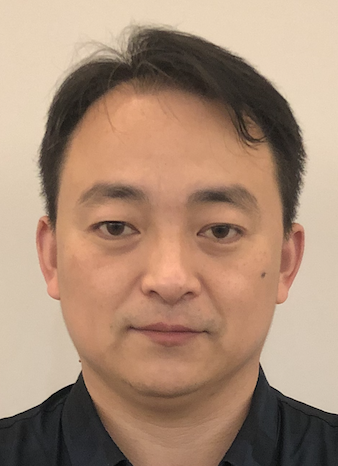}}]{Shanshe Wang}
  (Member, IEEE) received the B.S. degree from the Department of Mathematics, Heilongjiang University, Harbin, China, in 2004, the M.S. degree in computer software and theory from Northeast Petroleum University, Daqing, China, in 2010, and the Ph.D. degree in computer science from the Harbin Institute of Technology, Harbin, China, in 2014. He held a post-doctoral position with Peking University from 2016 to 2018. He joined the School of Electronics Engineering and Computer Science, Institute of Digital Media, Peking University, Beijing, where he is currently a Research Associate Professor. His current research interests include video compression and video quality assessment.
\end{IEEEbiography}

\begin{IEEEbiography}[{\includegraphics[width=1in,height=1.25in,clip,keepaspectratio]{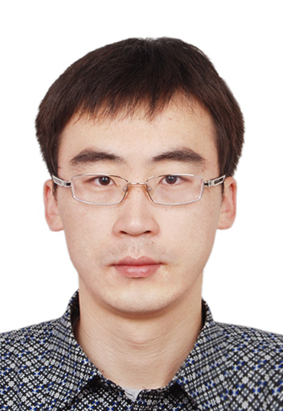}}]{Xinfeng Zhang}
  (Senior Member, IEEE) received the B.S. degree in computer science from the Hebei University of Technology, Tianjin, China, in 2007, and the Ph.D. degree in computer science from the Institute of Computing Technology, Chinese Academy of Sciences, Beijing, China, in 2014. From 2014 to 2017, he was a Research Fellow with the Rapid-Rich Object Search Laboratory, Nanyang Technological University, Singapore. From October 2017 to October 2018, he was a Postdoctoral Fellow with the School of Electrical Engineering System, University of Southern California, Los Angeles, CA, USA. From December 2018 to August 2019, he was a Research Fellow with the Department of Computer Science, City University of Hong Kong. He is currently an Associate Professor with the School of Computer Science and Technology, University of Chinese Academy of Sciences. He has authored more than 160 refereed journals/conference papers. He received the Best Paper Award of IEEE Multimedia 2018, the Best Paper Award at the 2017 Pacific-Rim Conference on Multimedia (PCM), and the Best Student Paper Award in IEEE International Conference on Image Processing 2018. His research interests include video compression and processing, image/video quality assessment, and 3D point cloud processing.
\end{IEEEbiography}

\begin{IEEEbiography}[{\includegraphics[width=1in,height=1.25in,clip,keepaspectratio]{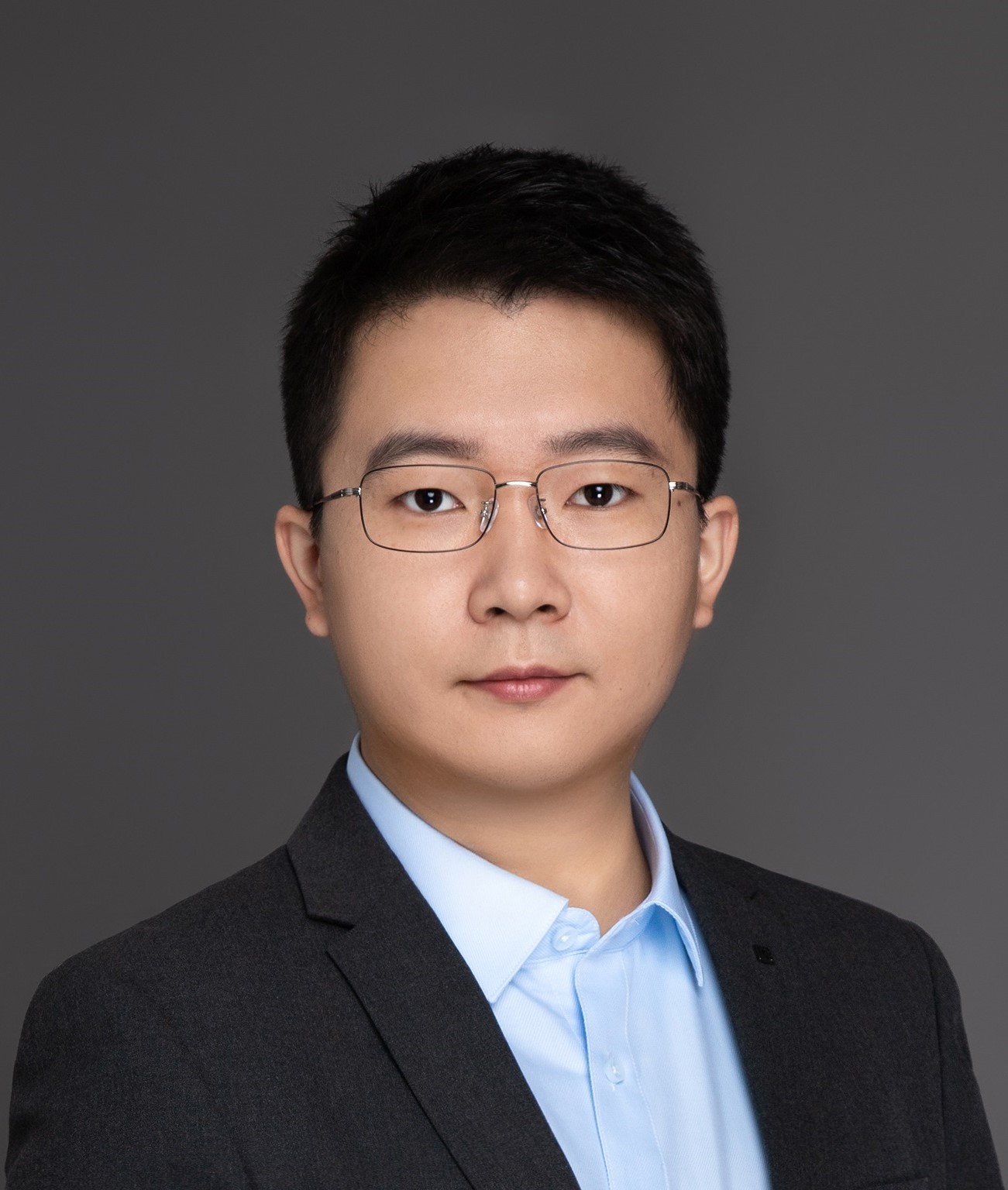}}]{Chuanmin Jia}
  (Member, IEEE) received the B.E. degree in computer science from the Beijing University of Posts and Telecommunications, Beijing, China, in 2015, and the Ph.D. degree in computer application technology from Peking University, Beijing, in 2020. He was a Visiting Student at New York University, USA, in 2018. He is currently an Assistant Professor with the Wangxuan Institute of Computer Technology, Peking University. His research interests include video compression and multimedia signal processing. He received the Best Paper Award of PCM 2017, the Best Paper Award of IEEE Multimedia 2018, and the Best Student Paper Award of IEEE MIPR 2019.
\end{IEEEbiography}

\begin{IEEEbiography}[{\includegraphics[width=1in,height=1.25in,clip,keepaspectratio]{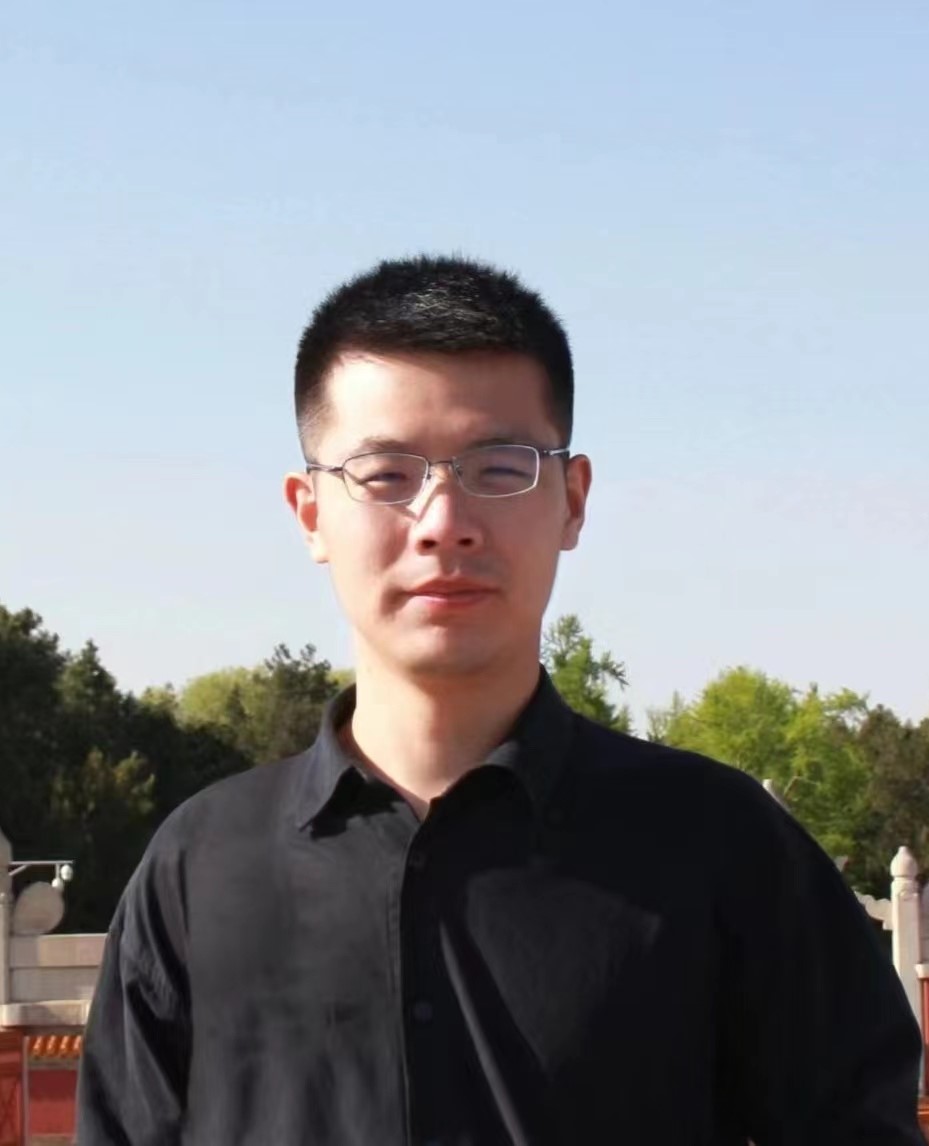}}]{Zhao Wang}
  (Member, IEEE) received the B.S. degree in software engineering from Fudan University in 2014, and the Ph.D. degree in computer science from Peking University in 2019. He previously worked at the DAMO Academy of Alibaba Group as an Algorithm Expert, and currently work at Peking University as an Engineer. His research interests include video compression, multi-modality vision coding and analysis. He has published more than 20 papers, and proposed more than 40 technical proposals to JVET, MPEG and AVS standards. He received the Best Student Paper Award from IEEE ICIP 2018, and his doctoral thesis was awarded the Excellent Doctoral Dissertation of Peking University.
\end{IEEEbiography}
  
\begin{IEEEbiography}[{\includegraphics[width=1in,height=1.25in,clip,keepaspectratio]{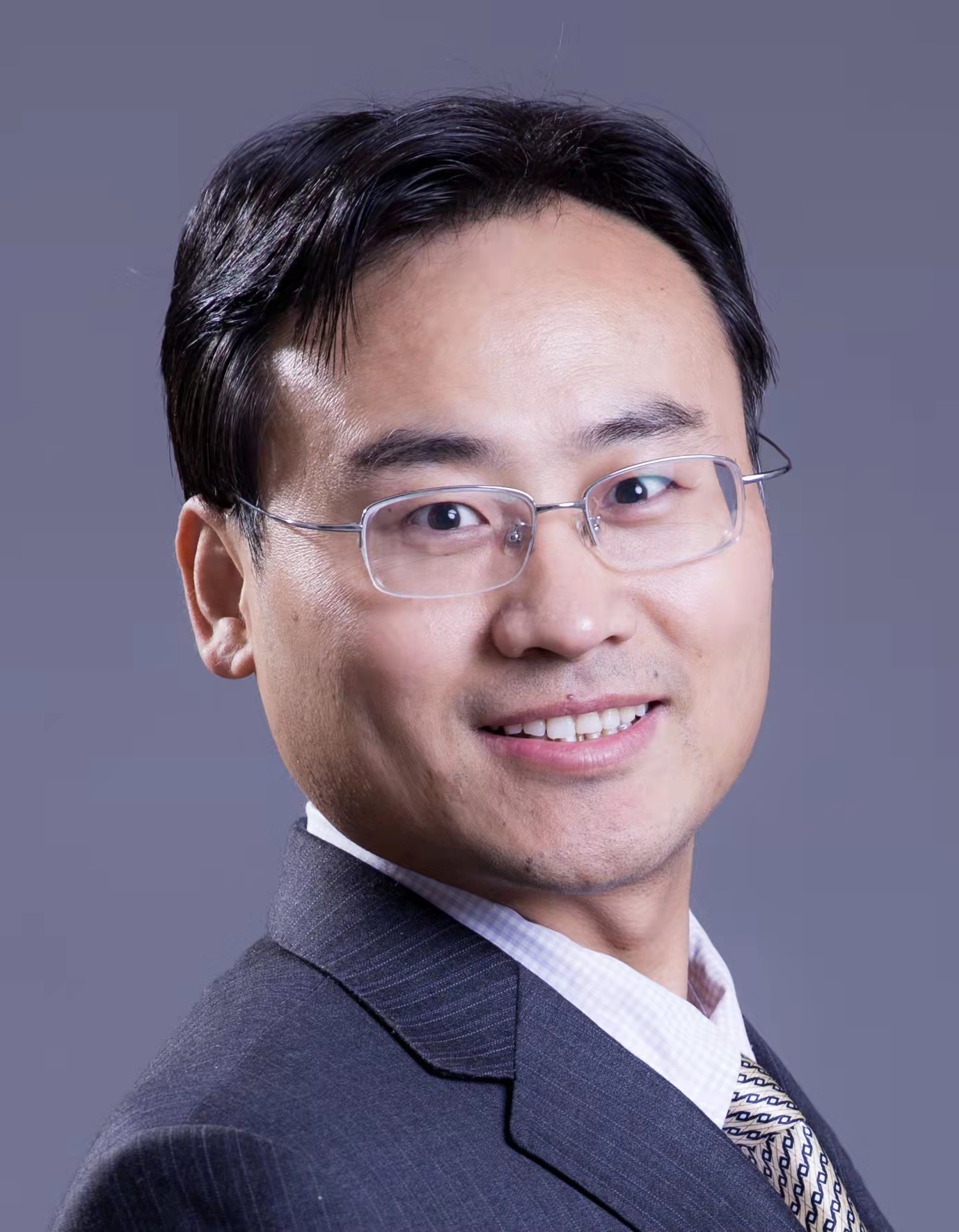}}]{Siwei Ma}
  (Fellow, IEEE) received the B.S. degree from Shandong Normal University, Jinan, China, in 1999, and the Ph.D. degree in computer science from the Institute of Computing Technology, Chinese Academy of Sciences, Beijing, China, in 2005. He held a postdoctoral position with the University of Southern California, Los Angeles, CA, USA, from 2005 to 2007. He joined the School of Electronics Engineering and Computer Science, Institute of Digital Media, Peking University, Beijing, where he is currently a Professor. He has authored over 300 technical articles in refereed journals and proceedings in image and video coding, video processing, video streaming, and transmission. He served/serves as an Associate Editor for the IEEE Transactions on Circuits and Systems for Video Technology and the Journal of Visual Communication and Image Representation.
\end{IEEEbiography}
  
\begin{IEEEbiography}[{\includegraphics[width=1in,height=1.25in,clip,keepaspectratio]{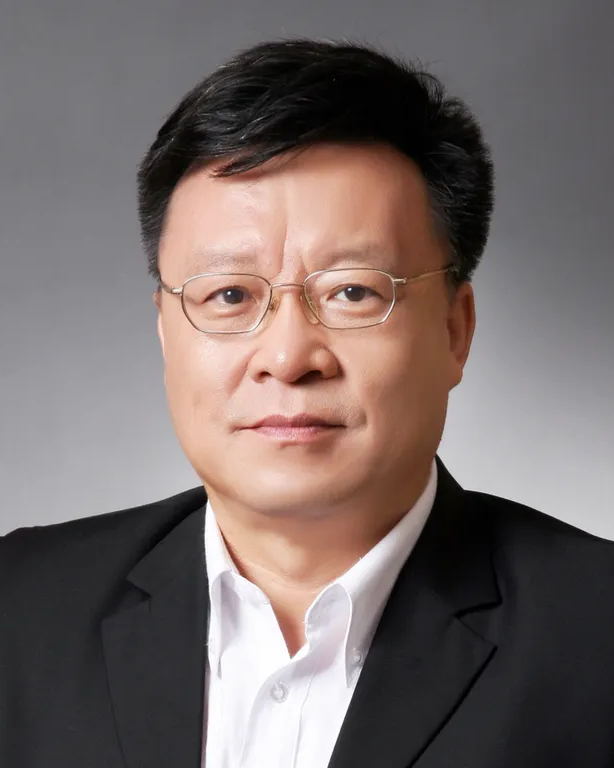}}]{Wen Gao}
  (Fellow, IEEE) received the Ph.D. degree in electronics engineering from The University of Tokyo, Japan, in 1991. He is currently a Boya Chair Professor in computer science at Peking University. He is also the Director of the Peng Cheng Laboratory, Shenzhen. Before joining Peking University, he was a Professor with the Harbin Institute of Technology, from 1991 to 1995. From 1996 to 2006, he was a Professor with the Institute of Computing Technology, Chinese Academy of Sciences. He has published extensively, including five books and over 1000 technical articles in refereed journals and conference proceedings in the areas of image processing, video coding and communication, computer vision, multimedia retrieval, multimodal interface, and bioinformatics. He served on the editorial boards for several journals, such as ACM CSUR, IEEE Transactions on Image Processing (TIP), IEEE Transactions on Circuits and Systems for Video Technology (TCSVT), and IEEE Transactions on Multimedia (TMM). He served on the advisory and technical committees for professional organizations. He was the Vice President of the National Natural Science Foundation (NSFC) of China, from 2013 to 2018, and the President of China Computer Federation (CCF), from 2016 to 2020. He is also the Deputy Director of China National Standardization Technical Committees. He is an Academician of the Chinese Academy of Engineering and a fellow of ACM.
\end{IEEEbiography}

\vfill

\include{appendix}

\end{document}

%% file: appendix.tex
\maketitle

\section*{Appendix}
\setcounter{table}{0}

\subsection{SMR Machine Library}

The information of machines in our SMR machine libraries for image classification and object detection is shown in TABLE~\ref{table:smr machine library for cls} and TABLE~\ref{table:smr machine library for det}, respectively. In TABLE~\ref{table:smr machine library for cls}, the leftmost column shows the library version. Note that v2 is built upon v1 so that it contains all machines from v1. The middle column shows the category of selected machines, and the rightmost column shows their names. In TABLE~\ref{table:smr machine library for det}, the first column shows the machines' names and the second column presents the neural network backbones that they use to extract features or their variants in different sizes.

\newpage

\bgroup
\def\arraystretch{1.28}
\begin{table*}[htbp]
    \centering
    \caption{SMR machine library for image classification.}
    \label{table:smr machine library for cls}
    \resizebox{\textwidth}{!}{
    \begin{tabular}{l|l|l}
    \Xhline{4\arrayrulewidth}
    Version & Category & Name \\ 
    \Xhline{3\arrayrulewidth}

    \multirow{13}{*}{v1} & \multirow{6}{*}{Classical CNN} & AlexNet \\ \cline{3-3} 

    &  & VGG-11, VGG-13, VGG-16, VGG-19, VGG-11-bn, VGG-13-bn, VGG-16-bn, VGG-19-bn \\ \cline{3-3} 

    &  & GoogLeNet; Inception-v3 \\ \cline{3-3} 

    &  & ResNet-18, ResNet-34, ResNet-50, ResNet-101, ResNet-152 \\ \cline{3-3} 

    &  & ResNeXt-50-32x4d, ResNeXt-101-32x8d; Wide-ResNet-50, Wide-ResNet-101 \\ \cline{3-3}

    &  & DenseNet-121, DenseNet-161, DenseNet-169, DenseNet-201 \\ \cline{2-3} 

    & \multirow{2}{*}{Lightweight Network} & ShuffleNet-v2-x0.5, ShuffleNet-v2-x1.0 \\ \cline{3-3} 

    &  & MobileNet-v2, MobileNet-v3-Small, MobileNet-v3-Large \\ \cline{2-3} 

    & \multirow{3}{*}{Network Architecture Search} & MNASNet-0.5, MNASNet-1.0 \\ \cline{3-3} 

    &  & EfficientNet-B0, EfficientNet-B1, EfficientNet-B3, EfficientNet-B5, EfficientNet-B7 \\ \cline{3-3} 

    &  & \begin{tabular}[c]{@{}l@{}}RegNet-x-400MF, RegNet-x-800MF, RegNet-x-1.6GF, RegNet-x-3.2GF, RegNet-x-8GF, RegNet-x-16GF, RegNet-x-32GF; \\ RegNet-y-400MF, RegNet-y-800MF, RegNet-y-1.6GF, RegNet-y-3.2GF, RegNet-y-8GF, RegNet-y-16GF, RegNet-y-32GF\end{tabular} \\ \cline{2-3} 

    & Transformer & ViT-b-16, ViT-b-32, ViT-l-16, ViT-l-32 \\ \cline{2-3} 

    & Modern CNN & ConvNeXt-Tiny, ConvNeXt-Small, ConvNeXt-Base, ConvNeXt-Large \\

    \Xhline{3\arrayrulewidth}

    \multirow{4}{*}{v2} &  Classical CNN & ResNeXt-101-64x4d \\ \cline{2-3} 

    & Lightweight Network & ShuffleNet-v2-x1.5, ShuffleNet-v2-x2.0 \\ \cline{2-3}

    & Network Architecture Search & \begin{tabular}[c]{@{}l@{}}MNASNet-0.75, MNASNet-1.3; \\ EfficientNet-B2, EfficientNet-B4, EfficientNet-B6; EfficientNet-v2-S, EfficientNet-v2-M, EfficientNet-v2-L\end{tabular} \\ \cline{2-3}

    & Transformer & Swin-T, Swin-S, Swin-B \\

    \Xhline{4\arrayrulewidth}
    \end{tabular}
    } 
\end{table*}
\egroup

\bgroup
\def\arraystretch{1.28}
\begin{table*}[htbp]
  \centering
  \caption{SMR machine library for object detection.}
  \label{table:smr machine library for det}
  \resizebox{0.8\textwidth}{!}{
  \begin{tabular}{l|l}
  \Xhline{4\arrayrulewidth}
  Method & Backbones/Variants \\
  \Xhline{3\arrayrulewidth}
  \multirow{4}{*}{Faster R-CNN} & ResNet-50, ResNet-101, ResNeXt-101-32x4d, ResNeXt-101-32x8d, ResNeXt-101-64x4d \\ \cline{2-2} 
   & ResNet-50-DCN, ResNet-101-DCN, ResNeXt-101-32x4d-DCN \\ \cline{2-2} 
   & Res2Net-101; ResNeSt-50, ResNeSt-101 \\ \cline{2-2} 
   & RegNet-x-400MF/800MF/1.6GF/3.2GF/4GF \\ \hline
  \multirow{4}{*}{RetinaNet} & ResNet-50, ResNet-101, ResNeXt-101-32x4d, ResNeXt-101-64x4d \\ \cline{2-2} 
   & ResNet-50-NAS-FPN \\ \cline{2-2} 
   & PVTv2-B0/B1/B2/B3/B4/B5 \\ \cline{2-2} 
   & RegNet-x-800MF/1.6GF/3.2GF \\ \hline
  \multirow{3}{*}{Cascade R-CNN} & ResNet-50, ResNet-101, ResNeXt-101-32x4d, ResNeXt-101-64x4d \\ \cline{2-2} 
   & ResNet-50-DCN, ResNet-101-DCN \\ \cline{2-2} 
   & Res2Net-101; ResNeSt-50, ResNeSt-101 \\ \hline
  SSD & SSD VGG-16-512; SSD-Lite MobileNet-v2 \\ \hline
  FCOS & ResNet-50, ResNet-101, ResNeXt-101-64x4d \\ \hline
  FoveaBox & ResNet-50, ResNet-101 \\ \hline
  RepPoints & ResNet-50, ResNet-101, ResNeXt-101-32x4d-DCN \\ \hline
  Grid R-CNN & ResNet-50, ResNet-101, ResNeXt-101-32x4d, ResNeXt-101-64x4d \\ \hline
  FSAF & ResNet-50, ResNet-101, ResNeXt-101-64x4d \\ \hline
  Generalized Focal Loss & ResNet-50, ResNet-101, ResNet-101-DCNv2, ResNeXt-101-32x4d, ResNeXt-101-32x4d-DCNv2 \\ \hline
  PAA & ResNet-50, ResNet-101 \\ \hline
  DETR & DETR ResNet-50; Deformable DETR ResNet-50; Deformable DETR ResNet-50 (refined) \\ \hline
  YOLOX & YOLOX-tiny/s/l/x \\ \hline
  YOLOv5 & YOLOv5-n/s/m/l-640, YOLOv5-n/s/m/l-1280 \\ \hline
  PP-YOLOE+ & PP-YOLOE+-s/m/l/x \\ \hline
  YOLOv7 & YOLOv7-tiny/l/x-640, YOLOv7-w/e-1280 \\ \hline
  YOLOv8 & YOLOv8-n/s/m/l/x \\ \hline
  \multirow{6}{*}{Others} & Dynamic R-CNN ResNet-50 \\ \cline{2-2} 
   & DetectoRS Cascade R-CNN ResNet-50 \\ \cline{2-2} 
   & CornerNet HourglassNet-104 \\ \cline{2-2} 
   & CentripetalNet HourglassNet-104 \\ \cline{2-2} 
   & AutoAssign ResNet-50 \\ \cline{2-2} 
   & DyHead ATSS ResNet-50 \\
   \Xhline{4\arrayrulewidth}
  \end{tabular}
  }
\end{table*}
\egroup